\documentclass[preprint,12pt]{elsarticle}

\usepackage{amsmath}
\usepackage{orcidlink}
\usepackage{graphicx} % Required for inserting images
\usepackage{xcolor,ulem}
\usepackage{multirow}
\usepackage{mathtools}
\usepackage{amsfonts}
\usepackage{amssymb}
\usepackage{amsthm}
\usepackage{booktabs}
\usepackage{algpseudocode}
\usepackage{hyperref}
\usepackage{natbib}
\usepackage{blindtext}
\usepackage{pdflscape} % For rotating the page
\usepackage{fancyhdr} % For custom headers and footers
\usepackage{float}
\usepackage{tikz}
\usepackage{natbib}
\usepackage{algorithm}
\usepackage{soul,cancel}

\usepackage{verbatim}
\usepackage{subcaption}
\usepackage{pdflscape}  % or \usepackage{lscape}

\usetikzlibrary{arrows.meta, positioning}
\graphicspath{ {./images/} }

\newcommand{\blue}{\color{blue}}
\newcommand{\red}{\color{red}}

\usepackage{soul}

\usetikzlibrary{positioning, arrows.meta}
\usetikzlibrary{arrows.meta, positioning, shapes.geometric}
\newtheorem{mydef}{Definition}

\journal{Information Sciences}

\begin{document}

\begin{frontmatter}



\title{Estimating Staged Event Tree Models via Hierarchical
Clustering on the Simplex}

 %% \affiliation[label1]{organization={}
%\affiliation[label2]{organization={}

\author[label1]{Muhammad Shoaib{\orcidlink{0009-0001-4064-0091}}} %% Author name
\author[label1]{Eva Riccomagno{\orcidlink{0000-0001-7280-2772}}}  
\author[label2]{Manuele Leonelli{\orcidlink{0000-0002-2562-5192}}}
\author[label3]{Gherardo Varando{\orcidlink{0000-0002-6708-1103}}}
\affiliation[label1]{organization={University of Genova},%Department and Organization
            addressline={Department of Mathematics}, 
            city={Genova},
            postcode={16146}, 
            %state={},
            country={Italy}}
\affiliation[label2]{organization={ IE University},%Department and Organization
            addressline={School of Science and Technology}, 
            city={Madrid},
            %postcode={16146}, 
            %state={},
            country={Spain}}
\affiliation[label3]{organization={Universitat de València},%Department and Organization
            addressline={Department of Statistics and Operational Research}, 
            city={València},
            postcode={}, 
            %state={},
            country={Spain}}

%% Abstract
\begin{abstract}
Staged tree models enhance Bayesian networks by incorporating context-specific dependencies through a stage-based structure. In this study, we present a new framework for estimating staged trees using hierarchical clustering on the probability simplex, utilizing simplex basesd  divergences. We conduct a thorough evaluation of several distance and divergence metrics including Total Variation, Hellinger, Fisher, and Kaniadakis—alongside various linkage methods such as Ward.D2, average, complete, and McQuitty. We conducted the simulation experiments that reveals Total Variation, especially when combined with Ward.D2 linkage, consistently produces staged trees with better model fit, structure recovery, and computational efficiency. We assess performance by utilizing relative Bayesian Information Criterion (BIC), and Hamming distance. Our findings indicate that although Backward Hill Climbing (BHC) delivers competitive outcomes, it incurs a significantly higher computational cost. On the other, Total Variation divergence with Ward.D2 linkage, achieves similar performance while providing significantly better computational efficiency, making it a more viable option for large-scale or time sensitive tasks.

\end{abstract}

%Graphical abstract

% %Research highlights
% \begin{highlights}
% \end{highlights}

%% Keywords
\begin{keyword}
Staged Event Trees, Hierarchical Clustering, Probability Simplex, Information Geometry


\end{keyword}

\end{frontmatter}



\section{Introduction}
\label{sec1}

Probabilistic graphical models (PGMs)~\cite{lauritzen1996graphical}, including Bayesian networks (BNs), offer a streamlined approach to represent multivariate distributions through relationships of conditional independence~\cite{nagarajan2014bayesian,scutari2021bayesian}. Although BNs are proficient at representing symmetric dependencies using directed acyclic graphs, they face challenges in modeling the context-specific asymmetries that are common in real-world decision-making processes. Staged event trees overcome this limitation by extending BNs into event trees featuring coloured nodes (referred to as \textit{stages}), with each colour denoting a distinct conditional probability distribution \cite{collazo2018chain, smith2008conditional}. These models maintain the interpretability of BNs while adeptly capturing heterogeneous dependencies, making them particularly suitable for applications in risk analysis \cite{varando2024staged}.

A key observation from \cite{smith2008conditional} is that every BN can be converted into a corresponding staged event tree, although not every staged event tree can be reduced to a BN on the observed variables.

This asymmetry underscores the greater expressive power of staged event trees, particularly in capturing localized or context-specific variations in dependency structures that standard BNs are not equipped to represent.

Staged event trees can be illustrated within the probability simplex because each complete path from the root to a leaf of the tree represents an atomic event in the sample space. Each path is given a probability that reflects the chance of that sequence of events happening. Together, all the paths in the tree create a probability mass function (PMF) over the atomic events, as the probabilities attributed to each root-to-leaf path total to one. Consequently, this PMF exists as a point in the probability simplex, which encompasses all valid probability distributions in the specified event space. This geometric perspective allows for the application of information-theoretic and information geometric methods to evaluate and contrast stages using divergence measures \cite{amari2016information, csiszar2004information}, thereby aiding in the clustering of similar conditional distributions of the same tree.

In this research, staged event trees group together non-leaf nodes (situations) that share the same conditional probability distributions concerning their possible future outcomes. Correctly identifying these stages is essential for the model's clarity and simplicity: having fewer stages results in a more easily understood model that preserves important context-specific structures while avoiding overfitting. Our framework tackles this issue by establishing a divergence metric for conditional distributions related to different nodes in the tree. For any chosen pair of nodes (contexts), we calculate different divergence methods for their estimated conditional probability distributions. This divergence measures how distinct the probabilistic behaviors are within these contexts. A low divergence implies that the contexts exhibit similar predictive behavior and can be grouped into the same stage; conversely, a high divergence indicates that they should remain distinct.

By calculating the divergence for all pairs of nodes qualified for staging, we generate a distance (or dissimilarity) matrix. This matrix functions as input for hierarchical clustering algorithms, which sequentially combine the most similar pairs of nodes (or clusters) according to their divergence until a predetermined stopping point is reached.

% We conduct a systematic evaluation of various traditional distances alongside information geometric and information theory divergences using four different linkage methods: average, Complete, McQuitty, and Ward's \cite{mcquitty1966similarity, ward1963hierarchical}. Our objective is to discover combinations that strike a balance between model simplicity and fidelity to the underlying data generating process.

To assess the effectiveness of clustering, we perform thorough simulation experiments using staged event tree models that differ in complexity and size. We utilize four assessment criteria: Bayesian Information Criterion (BIC) \cite{schwarz1978estimating}, and Hamming distance \cite{hamming1950error}.

We then apply the several divergence measures with the combination of ward.D2 method on several benchmark datasets. These datasets were used for supervised classification of categorical class variables $C$ given a vector of categorical features $(X_1,...,X_p)$, which can be found in Appendix \ref{tab:dataset_summary}.

%%%%%%%%%%%%%%%%%%%%%%%%%%%%
%%%%%%%%%%%%%%%%%%%%%%%%%%%%
%%%%%%%%%%%%%%%%%%%%%%%%%%%%

% ===== Section: Staged event trees and model definition =====
\section{Staged event trees and model definition}
\label{sec:stagedtrees}

We begin by fixing notation. For $[p]=\{1,\ldots,p\}$, let $X=(X_i)_{i\in[p]}$ denote a vector of discrete random variables, where each $X_i$ takes values in a finite state space $\mathcal X_i$. The joint sample space is $\mathcal X=\times_{i\in[p]}\mathcal X_i$. For any $A\subseteq[p]$, write $X_A=(X_i)_{i\in A}$ and, for a realization $x\in\mathcal X$, denote $x_A=(x_i)_{i\in A}\in\mathcal X_A=\times_{i\in A}\mathcal X_i$. We use $X_{-A}$ for the subvector indexed by $[p]\setminus A$. For $i\in[p]$, we set $\mathcal X_{[i]}=\times_{j\in[i]}\mathcal X_j$, the set of all partial assignments of the variables $(X_1,\ldots,X_i)$, that is, those preceding and including $X_i$ in the fixed order. 
%When needed, $P$ denotes the joint mass function on $\mathcal X$.

We work with event trees that unfold one variable at a time in a conceptual, not temporal, order: nodes encode partial assignments and edges append the next coordinate. Formally, the node at depth $i$ represents the history, i.e. the values of the conceptually preceding variables, $x_{[i]}=(x_1,\ldots,x_i)\in\mathcal X_{[i]}$. Each outgoing edge from $x_{[i]}$ is labeled by a value $x_{i+1}\in\mathcal X_{i+1}$ of $X_{i+1}$ and leads to the node $x_{[i+1]}=(x_{[i]},x_{i+1})$. 
%We call the ordered list of edge labels along a root--to--leaf path its \emph{label sequence}.

\begin{mydef}[$X$-compatible event tree]
\label{def:x-compatible-event-tree}
For a categorical random vector $X=(X_i)_{i\in[p]}$, an \textnormal{$X$-compatible event tree} is a rooted, directed tree $T=(V,E)$ with distinguished root $x_0\in V$ such that:
\begin{enumerate}
\item $V=\{x_0\}\cup\bigcup_{i\in[p]}\mathcal X_{[i]}$. A vertex $x_{[i]}\in\mathcal X_{[i]}$ represents the partial assignment $(x_1,\ldots,x_i)$ and has depth $i$.
\item For $i\in[p]$, an edge $(x_{[i-1]},x_{[i]})\in E$ exists if and only if $x_{[i-1]}\in\mathcal X_{[i-1]}$ and $x_{[i]}\in\mathcal X_{[i]}$ with $x_{[i]}=(x_{[i-1]},x_i)$.
\end{enumerate}
We write $E(v)=\{(v,w)\in E\}$ for the set of outgoing edges from a vertex $v$.
\end{mydef}

\begin{figure}[t]
\centering
\begin{subfigure}{0.49\textwidth}
\centering
\scalebox{0.6}{%
\begin{tikzpicture}[auto, scale=12,
Xwhite/.style={circle,inner sep=1mm,minimum size=0.4cm,draw,black,very thick,fill=white,text=black},
leaf/.style={circle,inner sep=1mm,minimum size=0.2cm,draw,very thick,black,fill=gray,text=black}
]
% --- global controls for edge labels ---
\newcommand{\edgelabeloffset}{0.5ex} % label–edge gap (increase to move further away)
\newcommand{\edgelabelpos}{0.62} % position along edge
\tikzset{
edgelabel/.style={pos=\edgelabelpos, sloped, font=\scriptsize},
elabove/.style={edgelabel, above, yshift=-\edgelabeloffset},
elbelow/.style={edgelabel, below, yshift=\edgelabeloffset},
}
% --- vertical spacing control ---
\newcommand{\yy}{1.60} % multiply all y-coordinates by this factor
% X positions
\def\xone{0.25} % depth 1
\def\xtwo{0.50} % depth 2
\def\xthree{0.75} % leaves
% Root
\node [Xwhite] (root) at (0.000, {0.500*\yy}) {};
% Depth 1 (white)
\node [Xwhite] (root-a) at (\xone, {0.850*\yy}) {};
\node [Xwhite] (root-b) at (\xone, {0.500*\yy}) {};
\node [Xwhite] (root-c) at (\xone, {0.150*\yy}) {};
% Depth 2 (white)
\node [Xwhite] (root-a-0) at (\xtwo, {0.920*\yy}) {};
\node [Xwhite] (root-a-1) at (\xtwo, {0.752*\yy}) {};
\node [Xwhite] (root-b-0) at (\xtwo, {0.584*\yy}) {};
\node [Xwhite] (root-b-1) at (\xtwo, {0.416*\yy}) {};
\node [Xwhite] (root-c-0) at (\xtwo, {0.248*\yy}) {};
\node [Xwhite] (root-c-1) at (\xtwo, {0.080*\yy}) {};
% Leaves (18, evenly spaced)
\node [leaf] (root-a-0-1) at (\xthree, {0.970000*\yy}) {};
\node [leaf] (root-a-0-2) at (\xthree, {0.914706*\yy}) {};
\node [leaf] (root-a-0-3) at (\xthree, {0.859412*\yy}) {};
\node [leaf] (root-a-1-1) at (\xthree, {0.804118*\yy}) {};
\node [leaf] (root-a-1-2) at (\xthree, {0.748824*\yy}) {};
\node [leaf] (root-a-1-3) at (\xthree, {0.693529*\yy}) {};
\node [leaf] (root-b-0-1) at (\xthree, {0.638235*\yy}) {};
\node [leaf] (root-b-0-2) at (\xthree, {0.582941*\yy}) {};
\node [leaf] (root-b-0-3) at (\xthree, {0.527647*\yy}) {};
\node [leaf] (root-b-1-1) at (\xthree, {0.472353*\yy}) {};
\node [leaf] (root-b-1-2) at (\xthree, {0.417059*\yy}) {};
\node [leaf] (root-b-1-3) at (\xthree, {0.361765*\yy}) {};
\node [leaf] (root-c-0-1) at (\xthree, {0.306471*\yy}) {};
\node [leaf] (root-c-0-2) at (\xthree, {0.251176*\yy}) {};
\node [leaf] (root-c-0-3) at (\xthree, {0.195882*\yy}) {};
\node [leaf] (root-c-1-1) at (\xthree, {0.140588*\yy}) {};
\node [leaf] (root-c-1-2) at (\xthree, {0.085294*\yy}) {};
\node [leaf] (root-c-1-3) at (\xthree, {0.030000*\yy}) {};
% Edges + labels (event tree: no probabilities)
\draw[->] (root) -- node [elabove] {$X_1=a$} (root-a);
\draw[->] (root) -- node [elabove] {$X_1=b$} (root-b);
\draw[->] (root) -- node [elbelow] {$X_1=c$} (root-c);
\draw[->] (root-a) -- node [elabove] {$X_2=0$} (root-a-0);
\draw[->] (root-a) -- node [elbelow] {$X_2=1$} (root-a-1);
\draw[->] (root-b) -- node [elabove] {$X_2=0$} (root-b-0);
\draw[->] (root-b) -- node [elbelow] {$X_2=1$} (root-b-1);
\draw[->] (root-c) -- node [elabove] {$X_2=0$} (root-c-0);
\draw[->] (root-c) -- node [elbelow] {$X_2=1$} (root-c-1);
\draw[->] (root-a-0) -- node [elabove] {$X_3=1$} (root-a-0-1);
\draw[->] (root-a-0) -- node [elabove] {$X_3=2$} (root-a-0-2);
\draw[->] (root-a-0) -- node [elbelow] {$X_3=3$} (root-a-0-3);
\draw[->] (root-a-1) -- node [elabove] {$X_3=1$} (root-a-1-1);
\draw[->] (root-a-1) -- node [elabove] {$X_3=2$} (root-a-1-2);
\draw[->] (root-a-1) -- node [elbelow] {$X_3=3$} (root-a-1-3);
\draw[->] (root-b-0) -- node [elabove] {$X_3=1$} (root-b-0-1);
\draw[->] (root-b-0) -- node [elabove] {$X_3=2$} (root-b-0-2);
\draw[->] (root-b-0) -- node [elbelow] {$X_3=3$} (root-b-0-3);
\draw[->] (root-b-1) -- node [elabove] {$X_3=1$} (root-b-1-1);
\draw[->] (root-b-1) -- node [elabove] {$X_3=2$} (root-b-1-2);
\draw[->] (root-b-1) -- node [elbelow] {$X_3=3$} (root-b-1-3);
\draw[->] (root-c-0) -- node [elabove] {$X_3=1$} (root-c-0-1);
\draw[->] (root-c-0) -- node [elabove] {$X_3=2$} (root-c-0-2);
\draw[->] (root-c-0) -- node [elbelow] {$X_3=3$} (root-c-0-3);
\draw[->] (root-c-1) -- node [elabove] {$X_3=1$} (root-c-1-1);
\draw[->] (root-c-1) -- node [elabove] {$X_3=2$} (root-c-1-2);
\draw[->] (root-c-1) -- node [elbelow] {$X_3=3$} (root-c-1-3);
\end{tikzpicture}%
}
\end{subfigure}
\hfill
\begin{subfigure}{0.49\textwidth}
\centering
\scalebox{0.6}{%
\begin{tikzpicture}[auto, scale=12,
Xroot/.style={circle,inner sep=1mm,minimum size=0.4cm,draw,black,very thick,fill=white,text=black},
Xgreen/.style={circle,inner sep=1mm,minimum size=0.4cm,draw,black,very thick,fill=green!60,text=black},
Xorange/.style={circle,inner sep=1mm,minimum size=0.4cm,draw,black,very thick,fill=orange!80,text=black},
Xcyan/.style={circle,inner sep=1mm,minimum size=0.4cm,draw,black,very thick,fill=cyan!50,text=black},
leaf/.style={circle,inner sep=1mm,minimum size=0.2cm,draw,very thick,black,fill=gray,text=black}
]
% same global controls
\newcommand{\edgelabeloffset}{0.5ex}
\newcommand{\edgelabelpos}{0.62}
\tikzset{
edgelabel/.style={pos=\edgelabelpos, sloped, font=\scriptsize},
elabove/.style={edgelabel, above, yshift=-\edgelabeloffset},
elbelow/.style={edgelabel, below, yshift=\edgelabeloffset},
}
% vertical spacing control
\newcommand{\yy}{1.60}
% X positions
\def\xone{0.25} % depth 1
\def\xtwo{0.50} % depth 2
\def\xthree{0.75} % leaves
% Root
\node [Xroot] (root) at (0.000, {0.500*\yy}) {};
% Depth 1 (green)
\node [Xgreen] (root-a) at (\xone, {0.850*\yy}) {};
\node [Xgreen] (root-b) at (\xone, {0.500*\yy}) {};
\node [Xgreen] (root-c) at (\xone, {0.150*\yy}) {};
% Depth 2 (top two orange, others cyan)
\node [Xorange] (root-a-0) at (\xtwo, {0.920*\yy}) {};
\node [Xorange] (root-a-1) at (\xtwo, {0.752*\yy}) {};
\node [Xcyan] (root-b-0) at (\xtwo, {0.584*\yy}) {};
\node [Xcyan] (root-b-1) at (\xtwo, {0.416*\yy}) {};
\node [Xcyan] (root-c-0) at (\xtwo, {0.248*\yy}) {};
\node [Xcyan] (root-c-1) at (\xtwo, {0.080*\yy}) {};
% Leaves (same coordinates as left)
\node [leaf] (root-a-0-1) at (\xthree, {0.970000*\yy}) {};
\node [leaf] (root-a-0-2) at (\xthree, {0.914706*\yy}) {};
\node [leaf] (root-a-0-3) at (\xthree, {0.859412*\yy}) {};
\node [leaf] (root-a-1-1) at (\xthree, {0.804118*\yy}) {};
\node [leaf] (root-a-1-2) at (\xthree, {0.748824*\yy}) {};
\node [leaf] (root-a-1-3) at (\xthree, {0.693529*\yy}) {};
\node [leaf] (root-b-0-1) at (\xthree, {0.638235*\yy}) {};
\node [leaf] (root-b-0-2) at (\xthree, {0.582941*\yy}) {};
\node [leaf] (root-b-0-3) at (\xthree, {0.527647*\yy}) {};
\node [leaf] (root-b-1-1) at (\xthree, {0.472353*\yy}) {};
\node [leaf] (root-b-1-2) at (\xthree, {0.417059*\yy}) {};
\node [leaf] (root-b-1-3) at (\xthree, {0.361765*\yy}) {};
\node [leaf] (root-c-0-1) at (\xthree, {0.306471*\yy}) {};
\node [leaf] (root-c-0-2) at (\xthree, {0.251176*\yy}) {};
\node [leaf] (root-c-0-3) at (\xthree, {0.195882*\yy}) {};
\node [leaf] (root-c-1-1) at (\xthree, {0.140588*\yy}) {};
\node [leaf] (root-c-1-2) at (\xthree, {0.085294*\yy}) {};
\node [leaf] (root-c-1-3) at (\xthree, {0.030000*\yy}) {};
% Edges + labels (with probabilities)
\draw[->] (root) -- node [elabove] {$X_1=a\ (0.3)$} (root-a);
\draw[->] (root) -- node [elabove] {$X_1=b\ (0.5)$} (root-b);
\draw[->] (root) -- node [elbelow] {$X_1=c\ (0.2)$} (root-c);
% Green stage: P(X2=0,1) = (0.6, 0.4)
\draw[->] (root-a) -- node [elabove] {$X_2=0\ (0.6)$} (root-a-0);
\draw[->] (root-a) -- node [elbelow] {$X_2=1\ (0.4)$} (root-a-1);
\draw[->] (root-b) -- node [elabove] {$X_2=0\ (0.6)$} (root-b-0);
\draw[->] (root-b) -- node [elbelow] {$X_2=1\ (0.4)$} (root-b-1);
\draw[->] (root-c) -- node [elabove] {$X_2=0\ (0.6)$} (root-c-0);
\draw[->] (root-c) -- node [elbelow] {$X_2=1\ (0.4)$} (root-c-1);
% Orange stage at depth 2: P(X3=1,2,3) = (0.7, 0.2, 0.1)
\draw[->] (root-a-0) -- node [elabove] {$X_3=1\ (0.7)$} (root-a-0-1);
\draw[->] (root-a-0) -- node [elabove] {$X_3=2\ (0.2)$} (root-a-0-2);
\draw[->] (root-a-0) -- node [elbelow] {$X_3=3\ (0.1)$} (root-a-0-3);
\draw[->] (root-a-1) -- node [elabove] {$X_3=1\ (0.7)$} (root-a-1-1);
\draw[->] (root-a-1) -- node [elabove] {$X_3=2\ (0.2)$} (root-a-1-2);
\draw[->] (root-a-1) -- node [elbelow] {$X_3=3\ (0.1)$} (root-a-1-3);
% Cyan stage at depth 2: P(X3=1,2,3) = (0.4, 0.4, 0.2)
\draw[->] (root-b-0) -- node [elabove] {$X_3=1\ (0.4)$} (root-b-0-1);
\draw[->] (root-b-0) -- node [elabove] {$X_3=2\ (0.4)$} (root-b-0-2);
\draw[->] (root-b-0) -- node [elbelow] {$X_3=3\ (0.2)$} (root-b-0-3);
\draw[->] (root-b-1) -- node [elabove] {$X_3=1\ (0.4)$} (root-b-1-1);
\draw[->] (root-b-1) -- node [elabove] {$X_3=2\ (0.4)$} (root-b-1-2);
\draw[->] (root-b-1) -- node [elbelow] {$X_3=3\ (0.2)$} (root-b-1-3);
\draw[->] (root-c-0) -- node [elabove] {$X_3=1\ (0.4)$} (root-c-0-1);
\draw[->] (root-c-0) -- node [elabove] {$X_3=2\ (0.4)$} (root-c-0-2);
\draw[->] (root-c-0) -- node [elbelow] {$X_3=3\ (0.2)$} (root-c-0-3);
\draw[->] (root-c-1) -- node [elabove] {$X_3=1\ (0.4)$} (root-c-1-1);
\draw[->] (root-c-1) -- node [elabove] {$X_3=2\ (0.4)$} (root-c-1-2);
\draw[->] (root-c-1) -- node [elbelow] {$X_3=3\ (0.2)$} (root-c-1-3);
\end{tikzpicture}%
}
\end{subfigure}
\caption{Event tree (left) and staged event tree (right) for $(X_1,X_2,X_3)$ with $\mathcal X_1={a,b,c}$, $\mathcal X_2={0,1}$, and $\mathcal X_3={1,2,3}$. Left: edges labeled as $X_i=\cdot$. Right: edges labeled as $X_i=\cdot$ with stage transition probabilities in parentheses.}
\label{fig:x123-side-by-side}
\end{figure}
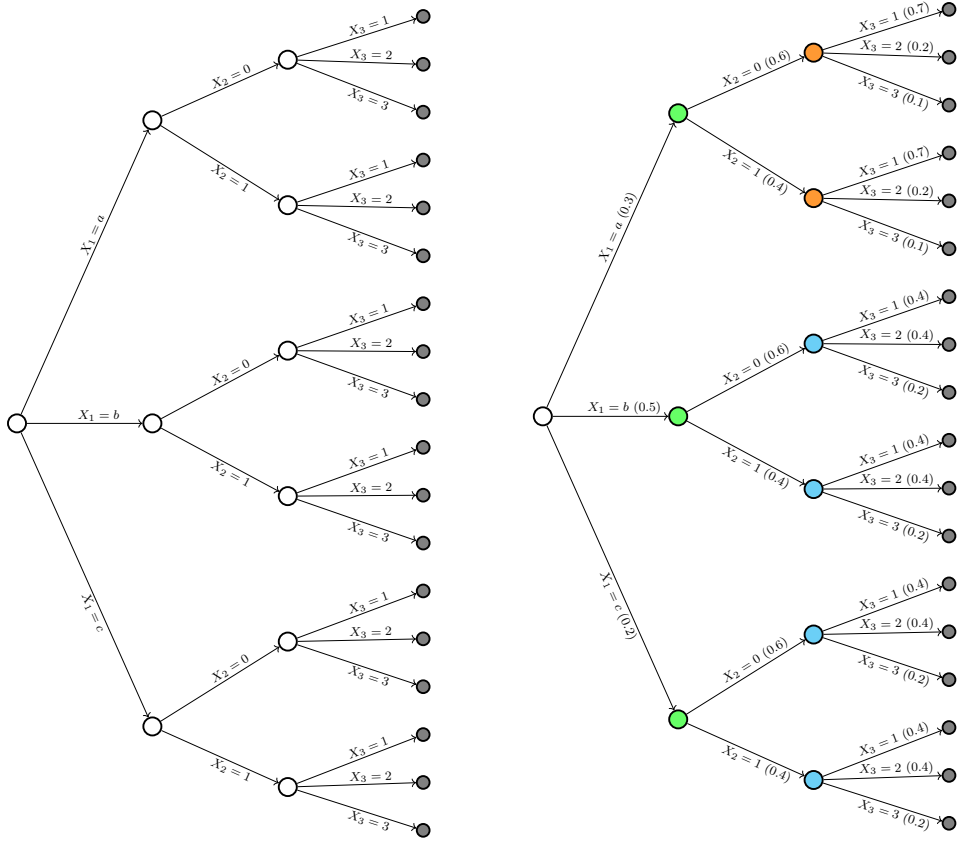

With this convention, the sequence of labels along any root--to--leaf path (equivalently, any leaf) encodes the unique full assignment $x=(x_1,\ldots,x_p)\in\mathcal X$. Figure~\ref{fig:x123-side-by-side} (left) illustrates the case $X=(X_1,X_2,X_3)$ with $\mathcal X_1=\{a,b,c\}$, $\mathcal X_2=\{0,1\}$, and $\mathcal X_3=\{1,2,3\}$; for example, the paths $a\!\to\!0\!\to\!3$ and $b\!\to\!1\!\to\!2$ correspond to the assignments $(a,0,3)$ and $(b,1,2)$. Depth-$i$ vertices are in one-to-one correspondence with the contexts $\mathcal X_{[i]}$; in particular, the six depth-2 vertices in Figure~\ref{fig:x123-side-by-side} (left) represent the histories $(x_1,x_2)\in\{a,b,c\}\times\{0,1\}$.

To pass from event trees to staged event trees, we \textit{embellish} an event tree by partitioning its internal vertices into \textbf{stages}. Two vertices are placed in the same stage when their outgoing edges admit a one-to-one correspondence that preserves labels (i.e., they have the same set of edge labels in the same order). Edge labels retain their meaning: the edge $(x_{[i-1]},x_{[i]})$ bears the newly appended symbol $x_i$. Only internal vertices are staged.

In a probabilistic reading, each edge carries a transition–probability parameter  $\theta_{\eta(e)}$. For any internal vertex at depth $i-1$, the parameters on its outgoing edges form a probability vector over $\mathcal{X}_i$ (positive and summing to one). Vertices placed in the same stage therefore share the same vector of outgoing parameters  (i.e. the same conditional distribution for the next variable) and this parameter tying across contexts is precisely how staged trees encode context-specific independences.

\begin{mydef}[$X$-compatible staged event tree]
\label{def:x-compatible-staged}
Let $T=(V,E)$ be an $X$-compatible event tree as in Definition~\ref{def:x-compatible-event-tree}. Fix a finite non-empty set $\mathcal C$ (the set of stages). An \textnormal{$X$-compatible staged event tree} is a triple $(T,\kappa,\eta)$ where:
\begin{enumerate}
\item $\kappa: \mathring{V} \to \mathcal C$ assigns a stage to each internal vertex, with $\mathring{V}=V\setminus\mathcal X_{[p]}$ {the set of non-leaf vertices}.
\item $\eta: E \to \mathcal C \times \bigcup_{i\in[p]}\mathcal X_i$ is the edge-labelling map given by
\[
\eta\bigl(x_{[i-1]},x_{[i]}\bigr)=\bigl(\kappa(x_{[i-1]}),\,x_i\bigr),\qquad i=1,\ldots,p.
\]
\end{enumerate}
Vertices $v,w\in \mathring{V}$ are in the same stage (i.e., $\kappa(v)=\kappa(w)$) if and only if their outgoing-edge labels coincide, equivalently $\eta(E(v))=\eta(E(w))$. Leaves are not staged.
\end{mydef}

Only internal vertices are staged: stages impose equality constraints on the distribution of the \textit{next} variable, which is undefined at leaves (no outgoing edges), so leaves are excluded. Moreover, the edge-labelling rule $\eta\bigl(x_{[i-1]},x_{[i]}\bigr)=(\kappa(x_{[i-1]}),x_i)$ ties the stage to the parent depth and to the value of $X_i$, ensuring that only vertices at the \textit{same depth} (hence governing the same variable $X_i$ and sharing the same sample space $\mathcal X_i$) can be placed in the same stage.

Figure~\ref{fig:x123-side-by-side} (right) embellishes the event tree on the left by staging, visualized via vertex colors. All depth-1 vertices share a single green stage, indicating that the conditional distribution of $X_2$ given $X_1$ is the same for all values of $X_1$; in words, $X_2$ is marginally independent of $X_1$. At depth 2, the vertices $(a,0)$ and $(a,1)$ form one orange stage, while the pairs $(b,0),(b,1)$ and $(c,0),(c,1)$ together form a second cyan stage. Within each stage the distribution of $X_3$ does not depend on $X_2$: the orange stage asserts this for $X_1=a$, and the cyan stage asserts it for $X_1\in\{b,c\}$. Because the orange and cyan stages differ, the behaviour of $X_3$ varies with $X_1$ but, once $X_1$ is fixed, not with $X_2$.

The staging in Definition~\ref{def:x-compatible-staged} induces a statistical model in the usual way. 
For each internal vertex \(v\) at depth \(i-1\), attach a simplex-valued parameter vector over the next variable \(X_i\); equivalently, to each outgoing edge \(e\) from \(v\) assign a parameter \(\theta_{\eta(e)}\in(0,1)\) with \(\sum_{f\in E(v)} \theta_{\eta(f)}=1\). 
Vertices in the same stage share these parameters because \(\eta(e)\) depends only on the stage of \(v\) and the next value. 
For a leaf \(l\), let \(\lambda(l)\) be the unique root--to--leaf path; its joint mass is the product of the edge parameters along \(\lambda(l)\). 
%This parameter sharing is exactly how staged trees encode context-specific independences.

\begin{mydef}[Staged event tree model]\label{def:set-model}
Let $(T,\kappa,\eta)$ be an $X$-compatible staged event tree as in Definition~\ref{def:x-compatible-staged}. 
Let $\mathfrak l_T$ be the set of leaves and, for $l\in\mathfrak l_T$, let $\lambda(l)$ denote the unique root--to--leaf path.
The parameter space is
\[
\Theta_T \;=\; \Bigl\{ \theta=(\theta_e)_{e\in\eta(E)} \in (0,1)^{|\eta(E)|} \ :\ 
\sum_{f\in E(v)} \theta_{\eta(f)} = 1 \ \text{for every } v\in V^{\circ} \Bigr\}.
\]
Define
\[
\varphi_T:\Theta_T\longrightarrow \Delta^{\circ}_{|\mathfrak l_T|-1}, 
\qquad 
\theta \longmapsto \Bigl( \prod_{f\in E(\lambda(l))} \theta_{\eta(f)} \Bigr)_{l\in\mathfrak l_T},
\]
where $\Delta^{\circ}_{m-1}$ denotes the open $(m\!-\!1)$-simplex.
The \textnormal{staged event tree model} is $\mathcal M_T=\mathrm{im}(\varphi_T)$.
\end{mydef}

By Definition~\ref{def:set-model}, the probability assigned to any full assignment is the product of the edge probabilities along the corresponding root--to--leaf path. Accordingly, Table~\ref{tab:x123-joint-products} lists all atoms for Figure~\ref{fig:x123-side-by-side} (right), each written as a product of stage-specific transitions.

Every Bayesian network model on $X$ admits an $X$-compatible staged event tree representation  \cite{smith2008conditional}, with explicit constructions that transform a BN into a staged tree given by \cite{varando2024staged}.  The converse does not hold: staging allows equality constraints among conditional distributions that vary by context, yielding context-specific independences and asymmetric dependence patterns  \cite{varando2024staged}. In contrast, DAG-based BNs encode only global, context-free conditional independences via (the symmetric) absence of edges, so some staged trees have no BN equivalent.

\begin{table}[t]
\centering
\caption{Joint distribution induced by the staged tree in Figure~\ref{fig:x123-side-by-side} (right). }
\label{tab:x123-joint-products}
\scalebox{0.8}{
\begin{tabular}{cccc @{\hskip 1.2cm} cccc}
\toprule
$X_1$ & $X_2$ & $X_3$ & $P(X_1,X_2,X_3)$
& $X_1$ & $X_2$ & $X_3$ & $P(X_1,X_2,X_3)$ \\
\midrule
% -------- LEFT BLOCK (9 rows): all a (6) + b with X2=0 (3)
$a$ & $0$ & $1$ & $0.3\times0.6\times0.7=0.1260$
& $b$ & $1$ & $1$ & $0.5\times0.4\times0.4=0.0800$ \\
$a$ & $0$ & $2$ & $0.3\times0.6\times0.2=0.0360$
& $b$ & $1$ & $2$ & $0.5\times0.4\times0.4=0.0800$ \\
$a$ & $0$ & $3$ & $0.3\times0.6\times0.1=0.0180$
& $b$ & $1$ & $3$ & $0.5\times0.4\times0.2=0.0400$ \\
$a$ & $1$ & $1$ & $0.3\times0.4\times0.7=0.0840$
& $c$ & $0$ & $1$ & $0.2\times0.6\times0.4=0.0480$\\
$a$ & $1$ & $2$ & $0.3\times0.4\times0.2=0.0240$
& $c$ & $0$ & $2$ & $0.2\times0.6\times0.4=0.0480$ \\
$a$ & $1$ & $3$ & $0.3\times0.4\times0.1=0.0120$
&$c$ & $0$ & $3$ & $0.2\times0.6\times0.2=0.0240$  \\
$b$ & $0$ & $1$ & $0.5\times0.6\times0.4=0.1200$ & $c$ & $1$ & $1$ & $0.2\times0.4\times0.4=0.0320$ 
 \\
$b$ & $0$ & $2$ & $0.5\times0.6\times0.4=0.1200$
& $c$ & $1$ & $2$ & $0.2\times0.4\times0.4=0.0320$ \\
$b$ & $0$ & $3$ & $0.5\times0.6\times0.2=0.0600$
& $c$ & $1$ & $3$ & $0.2\times0.4\times0.2=0.0160$ \\
\bottomrule
\end{tabular}}
\end{table}

Although we focus on $X$-compatible staged trees \citep[both because they are widely used and because efficient learning procedures are available,][]{JSSv102i06}, staged tree formalisms that are not $X$-compatible exist \cite{leonelli2019,carter2024pgm}.

\subsection{Learning staged event trees}

Learning the structure of a staged event tree from data is feasible but nontrivial, because the search space grows rapidly with the number of variables and with the sizes of their state spaces. A widely used approach is agglomerative hierarchical clustering, basically a backward hill climbing algorithm (BHC)~\citep{freeman2011bayesian}: start from the finest staging (each internal vertex in its own stage) and iteratively merge the pair of stages that yields the largest improvement in a chosen score, stopping when no merge helps. The original work employed a Bayesian score; subsequent implementations support alternative criteria such as BIC and AIC \citep{JSSv102i06}. More recent methods restrict the search to curated families of staged trees or address robustness and scalability, e.g., \citep{carli2023new,leonelli2022highly,leonelli2024structural,rios2024scalable}.

Given a dataset $\mathcal{D}$ of $n$ observations of the categorical vector $X$, the first step is to construct the \textit{saturated} $X$-compatible event tree, in which every internal vertex corresponds to a unique context (partial assignment) and therefore to a distinct conditional distribution. Each outgoing edge $e$ from a vertex $v$ at depth $i-1$ represents a possible transition to a value $x_i \in \mathcal{X}_i$ of the next variable. The associated edge probability parameter is denoted by $\theta_{\eta(e)}$, and fitting the model means estimating these parameters from the data in $\mathcal{D}$. Specifically, let $n(x_{[i-1]},x_i)$ denote the observed frequency of the transition $(x_{[i-1]},x_i)$ in $\mathcal{D}$. The fitted conditional probability of $X_i$ given the context $x_{[i-1]}$ is then
\[
\hat{\theta}_{\eta(e)} 
= \hat{P}(X_i=x_i \mid X_{[i-1]}=x_{[i-1]})
= \frac{n(x_{[i-1]},x_i) + \alpha}
       {\sum_{x_i' \in \mathcal{X}_i} n(x_{[i-1]},x_i') + \alpha|\mathcal{X}_i|},
\,\,\, x_i \in \mathcal{X}_i,
\]
where $\alpha$ is a smoothing parameter ($\alpha=0$ for MLE, $\alpha=1$ for Laplace, $\alpha=\tfrac12$ for Jeffreys). The collection of these estimated edge probabilities $\{\hat{\theta}_{\eta(e)}\}$ fully parameterizes a \textit{fitted} saturated staged event tree, which represents the empirical joint distribution $\hat{P}(X_1,\ldots,X_p)$ obtained from $\mathcal{D}$.

Learning an $X$-compatible staged event tree then amounts to estimating a \textit{coarser} stage partition that groups vertices exhibiting similar conditional behavior. Each internal vertex $v$ can be represented by its estimated conditional probability vector
$
\hat{\theta}_v = \bigl(\hat{\theta}_{\eta(e)} : e \in E(v)\bigr),$  
viewed as a point in the simplex $\Delta_{|\mathcal{X}_i|-1}$. The goal of structure learning is to cluster these vectors using a suitable distance or divergence measure so that vertices within the same cluster (stage) share similar conditional distributions. After clustering, the stage parameters are refitted by pooling counts across all vertices in the same cluster and recomputing the corresponding conditional probabilities.

In the following section, we review methods for hierarchical clustering on the space of probability vectors. We explore distances and divergences on the simplex, which allow hierarchical clustering algorithms to be applied directly to the empirical frequency vectors attached to each internal vertex.

\section{Hierarchical clustering on the simplex}
\label{Clus}

Unlike unrestricted Euclidean spaces, the simplex is a bounded domain defined by positivity and sum-to-one constraints. As a result, conventional vector space operations are not directly applicable for clustering within the simplex. 
%Instead, a non-Euclidean (curved) geometry can be defined to account for its unique structure \cite{amari2000methods, pistone2024unified}. 
Clustering aims to group probability distributions that share similar characteristics, while respecting the compositional nature of the simplex. This scenario frequently emerges in various statistical and machine learning contexts, including latent class models \cite{goodman1974exploratory}, multinomial data \cite{bishop1975discrete}, and structured probabilistic models \cite{lauritzen1996graphical}.
Effective clustering methods on the simplex often require rethinking fundamental concepts (such as distance, variability, and averaging) in a manner that is consistent with its intrinsic geometry \cite{amari2016information,aitchison1982statistical, pawlowsky2015modeling}.

We develop methods to cluster stages in staged event trees, grouping contexts that share similar conditional distributions so as to uncover the model’s conditional independence structure. Our approach begins by revisiting distances and divergences on the probability simplex for comparing stage-wise categorical distributions.

\subsection{ 
Distance metrics and divergences on the simplex
}
\label{section:2}

% \textcolor{red}{Also here some examples would be nice, we could reuse the Example in figure \ref{fig:example}. We could plot a 3 dimensional simplex and the conditional probabilities of $X_3$ from the example.}

% \begin{figure}[H]
%     \centering
%     \includegraphics[width=0.5\textwidth]{images/ternary_plot.pdf}
%    \caption{Illustration of a 3-dimensional simplex with conditional probabilities of $X_3$ derived from the staged event tree (right panel) in Figure~\ref{fig:x123-side-by-side}.}
%     \label{fig:ternary}
% \end{figure}

% The probability simplex provides an explicit geometric representation of the space of all probability distributions over a finite set of outcomes. This space is denoted by \( \Delta^{d-1} \) where \( d \) is the cardinality of $\mathcal X$ and the probability simplex $\Delta^{d-1}$ consists of all vectors $q=(q_1,\ldots,q_d)$ such that $q_i \geq 0$
% for all $i$ and $\sum_{i=1}^d q_i=1$. 
% It is a convex set because the line segment connecting any of its points \( p \) and \( q \), 
% \begin{equation*}
%     [0,1] \ni t \mapsto p(t) = (1 - t)p + t q,
% \end{equation*}
% lies entirely within \( \Delta^{d-1} \) \cite{pistone2024unified}.

A staged event tree model is a set of a probability distributions satisfying the same collection of conditional probability independences embedded in the stage structure of the tree. 
%and thus each element of the stage tree models can be represented as a point in \( \Delta^{d-1} \). 
%For any such probability distribution in the event tree model, 
A leaf node in the event tree represents an atomic event 
whose probability is derived from the product of the conditional probabilities along the path from the root to the leaf. 
When all transition probabilities are defined and known, the staged event tree encapsulates exactly one distribution in \( \Delta_{s-1} \), where $s=|\mathbb{X}|$ is the size of sample space. 
However, when one or more atomic event probabilities or the transition probabilities along the edges of the tree are unknown or parameterized, the staged event tree denotes a statistical model that encompasses a variety of probability distributions rather than a singular point in \( \Delta_{s-1} \). 

%%A staged event tree can be {\blue seen} \st{represented} as a point in \( \Delta^{d-1} \), \st{as the} {\blue and it represents a}  collection of conditional probability distributions \st{it defines uniquely correspond to a point within the simplex}. Each atomic event is linked to a leaf node of the tree, and its probability is derived from the product of the conditional probabilities along the path from the root to the leaf. Therefore, the entire set of atomic event probabilities defined by the staged event tree constitutes a valid probability vector in \( \Delta^{d-1} \). When all transition probabilities are defined and known, the staged event tree encapsulates a single distribution. 

In this second scenario, the model constitutes a submanifold or, under certain structural conditions, a curve embedded within \( \Delta_{s-1} \). This geometric viewpoint has been examined 
%in the research 
by \cite{leonelli2022curved}, who demonstrated that with appropriate regularity conditions, the statistical model linked to a staged event tree can be classified as a curved exponential family of the second order. This viewpoint indicates that staged event tree models can incorporate a differential-geometric structure typical of information geometry~\cite{amari2016information,pistone2013nonparametric, pistone2019information}. Specifically, the conditional probability distributions defined along the branches of the tree can be examined using notions like affine connections and divergence functions within the probability simplex.  
Viewing staged event trees through exponential and mixture representations, which align with dual affine coordinate systems, makes their connection to information geometry explicit \citep{pistone2019information}.
%By viewing the model through the lens of exponential and mixture representations—which align with dual affine coordinate systems—a formal link is established between staged event trees and the information geometry framework has been discussed in~\cite{pistone2019information} among others.
This relationship allows for the application of geometric methods, such as divergence-based distances and natural gradients, in the analysis, learning, and comparison of staged event tree models. 
We cluster vertices by comparing their conditional distributions. The comparison uses proper distances on $\Delta_{s-1}$ and divergences from information geometry. % \cite{amari2016information}.  
The measures we use are  implemented in the  \texttt{stagedtrees}  R package~\cite{JSSv102i06}, and are described below.

\begin{table}[t]
\small
\renewcommand{\arraystretch}{1.2} 
\centering
\caption{Distances used in Section~\ref{sec:simulation}.}
\label{tab:distance_functions}
\begin{tabular}{l l l}
\toprule
Distance & Formula & Range \\
\midrule
Total Variation distance & 
$d_{TV}(p,q) = \frac{1}{2} \sum_{i=1}^s |p_i - q_i|$ & 
$[0, 1]$ \\ 
Hellinger distance & 
$d_{H}(p,q) = \frac{1}{\sqrt{2}} \sqrt{\sum_{i=1}^s \left( \sqrt{p_i} - \sqrt{q_i} \right)^2}$ & 
$[0, 1]$ \\ 
% \footnote{not used}
% Euclidean distance & 
% $\displaystyle d_{E}(p,q) = \sqrt{\sum_{i=1}^d (p_i - q_i)^2}$ & 
% $[0, +\infty)$ \\
\bottomrule
\end{tabular}
\end{table}

\subsubsection{Distance metrics}
We focus on distance functions `${dist}$' that satisfy the standard axioms of a metric on the probability simplex $\Delta_{s-1}$; that is, for all $p,q\in \Delta_{s-1}$ the following four conditions hold:

\begin{enumerate}
    \item {non-negativity}:
    \(
    {dist}(p, q) \geq 0
    \)
    \item {identity of indiscernibles}:
    \(
    {dist}(p, q) = 0 \iff p = q
    \)
    \item {symmetry}:
    \(
    {dist}(p, q) = {dist}(q, p)
    \)
    \item {triangle inequality}:
    \(
    {dist}(p, r) \leq {dist}(p, q) + {dist}(q, r)
    \)
\end{enumerate}

Table~\ref{tab:distance_functions} lists the two distance functions used in Section~\ref{sec:simulation}, along with the respective ranges of values they attain. 
 Experiments with other distances yielded similar results and are therefore not discussed further. 
The \textit{Total Variation Distance}, $d_{TVD}$, is specifically designed for probability distributions and inherently respects the geometry of the simplex. The \textit{Hellinger} distance, $d_{H}$, is of particular interest, as it belongs to the class of Bregman divergences~\cite{bregman1967relaxation}  and maintains a strong connection to the geometry of the probability simplex.

\subsubsection{Divergences on the simplex}
In contrast to distances, divergences are generally asymmetric and need not satisfy the triangle inequality. However, they offer greater flexibility in representing directional differences between distributions. Divergences grounded in information geometry often respect the intrinsic curvature of the probability simplex, making them particularly well suited for probabilistic modeling. When symmetry is required for clustering, we use the symmetrized form (e.g., $d^{\mathrm{sym}}(p,q)=d(p\Vert q)+d(q\Vert p)$ for any divergence function $d$). In this paper, the divergences in Table~\ref{tab:divergence_functions} (Fisher in its squared form, Jensen--Shannon, symmetrized Kaniadakis, and total KL/Jeffreys) are nonnegative, equal zero if and only if $p=q$, and are symmetric (by definition where needed), but they do not satisfy the triangle inequality and therefore are not metrics on $\Delta_{s-1}$. If a true metric is needed, one can use the unsquared Fisher--Rao distance or the square root of the Jensen--Shannon divergence.

\begin{table}[t]
\small
\caption{Divergences used in Section~\ref{sec:simulation}.}
\label{tab:divergence_functions}
\centering
\renewcommand{\arraystretch}{1.2}
\scalebox{0.91}{
\begin{tabular}{l l l}
\toprule
Divergence & Formula & Range \\
\midrule
Fisher divergence &
$\displaystyle d_F(p\Vert q) = 4\,\arccos^{2}\!\left(\sum_{i=1}^{s} \sqrt{p_i\,q_i}\right)$ &
$\left[0, \frac{\pi}{2}\right]$ \\
Jensen--Shannon divergence &
$\displaystyle d_{\mathrm{JS}}(p\Vert q) = \tfrac{1}{2} \sum_{i=1}^{s}
\!\left[
p_i \log\!\frac{2p_i}{p_i+q_i}
+ q_i \log\!\frac{2q_i}{p_i+q_i}
\right]$ &
$[0, \ln 2]$ \\
Kaniadakis divergence &
$\displaystyle {d}_\kappa(p \,\|\, q) = \sum_{x \in \Omega} \left( \log_\kappa(p(x)) - \log_\kappa(q(x)) \right) \cdot \frac{A(p(x))}{\sum_{x \in \Omega} A(p(x))}$ &
$[0, +\infty)$ \\
Total KL divergence &
$\displaystyle d_{T}(p\Vert q) = \sum_{i=1}^{s}
\left(
p_i \log\!\frac{p_i}{q_i}
+ q_i \log\!\frac{q_i}{p_i}
\right)$ &
$[0, +\infty)$ \\
\bottomrule
\end{tabular}}
\end{table}

A central quantity in the Fisher divergence is the Bhattacharyya coefficient 
$\sum_{i=1}^s \sqrt{p_i q_i}$. 
It links to the Hellinger distance via 
$d_H^2(p,q)=2\bigl(1-\cos\bigl(\tfrac{d_F(p,q)}{2}\bigr)\bigr)$
and to the Riemannian geometry of the simplex~\citep{amari2016information,bhattacharyya1943measure}.
Applications to compositional data appear in~\citep{erb2021information}.

%The Fisher divergence \st{, rooted in Information Geometry,} incorporates the Bhattacharyya coefficient $\sum_{i=1}^d \sqrt{p_i q_i}$  as a central component~\cite{amari2016information, bhattacharyya1943measure}. This divergence is closely related to the Hellinger distance, as both are based on square-root transformations of probability distributions and reflect the same underlying Riemannian geometry of the simplex~\cite{amari2016information}. Applications of Fisher divergence to compositional data have been explored in~\cite{erb2021information}.

The Jensen–Shannon divergence is a symmetrized and smoothed variant of the KL divergence that remains finite even when the support of 
$q$ does not fully contain that of $p$~\cite{lin1991divergence}; in other words, it does not require 
$q_i>0$
wherever 
$p_i>0$, making it more robust in sparse or degenerate cases.

%The Jensen-Shannon divergence represents a symmetrized and smoothed adaptation of the KL divergence that remains finite even when the support of \( q \) does not encompass that of \( p \) \cite{lin1991divergence}. 

The Kaniadakis divergence modifies the classical logarithmic transformation by employing generalized logarithms and escort probabilities, offering a more flexible framework for measuring divergence in non-standard probabilistic settings~\cite{pistone2023kaniadakis, naudts2008generalised}.

The total KL divergence is defined as the sum of the KL divergences in both directions between two distributions. It is nonnegative, vanishes only at equality, and is symmetric, though it is not a metric since the triangle inequality does not hold \citep{amari2016information}. 

% \st{We do not use the alternative convention that sums KLs to a common reference.}\footnote{\blue{The sentence "We do not... to a common reference"} does not contributo to the economy of the paper and it is subject to different interpretation. Hence I suggest to remove it.}

%The total KL divergence can be interpreted as the sum of the KL divergences between a distribution $p$ with respect to a distribution $q$, where the KL divergence is defined as 
%$d_{KL}(p\| q) = \sum_{i=1}^d \left(p_i \log \frac{p_i}{q_i} \right)$; it can also be interpreted as a particular metric that quantifies the overall difference between two distributions. When comparing multiple probability distributions, the total Kullback–Leibler (KL) divergence is typically defined as the sum of the individual KL divergences between a chosen reference distribution and each distribution in the collection.~\cite{amari2016information}. 

% Lastly, Total KL divergence is characterised by the symmetric sum of KL divergences calculated in both directions \cite{amari2016information}.\footnote{\colour{blue}{It does not seem correct to me. Please check.}}

%%%%%%%%%%%%%%%%%%%%%%%%%%%%
%%%%%%%%%%%%%%%%%%%%%%%%%%%%
%%%%%%%%%%%%%%%%%%%%%%%%%%%%

\subsection{Hierarchical clustering for $X$-compatible staged event trees}
\label{sec:sevtHC}

Hierarchical clustering can be adapted to the probability simplex, where Euclidean assumptions do not apply. For \(X\)-compatible staged event trees, we cluster the stage probability vectors using a chosen distance or divergence on \(\Delta_{s-1}\). The procedure iteratively merges stages while preserving probabilistic structure, yielding increasingly coarse approximations of the original tree. The resulting dendrogram reveals multiscale relationships among contexts and supports model selection, regularization, and interpretation. This geometry-aware approach aligns with structural learning objectives in probabilistic graphical models and helps maintain semantic coherence in staged event trees \citep{JSSv102i06,amari2000methods,pistone2024unified}. We next present the algorithm and the specific clustering criteria used in Section~\ref{sec:simulation}.

\begin{algorithm}
\caption{Learning Stage Structure via Hierarchical Clustering}
\label{alg:hc-stages}
\begin{algorithmic}[1]
\Require Dataset $\mathcal{D}$ on $X=(X_1,\dots,X_p)$; symmetric dissimilarity $d$ on the simplex; linkage $\mathsf{link}$; model score $S(\cdot)$ (e.g., BIC);
number of stages $\{k_i\}_{i=2}^p$ (optional)

\State Initialise the saturated $X$-compatible event tree $T$, where each internal vertex forms its own stage.
\For{$i=2,\dots,p$}
  \State Let $\mathcal{S}_i$ be the set of internal vertices at depth $i-1$, each with outgoing edge-labels in  $\mathcal{X}_i$.
  \State Using $\mathcal{D}$, compute for each $v\in\mathcal{S}_i$ the (optionally smoothed) empirical conditional vector $\hat{\theta}_v\in\Delta_{|\mathcal{X}_i|-1}$.
  \State Build the dissimilarity matrix $M_i=\bigl[d(\hat{\theta}_v,\hat{\theta}_w)\bigr]_{v,w\in\mathcal{S}_i}$.
  \State Apply hierarchical agglomerative clustering on $M_i$ with linkage $\mathsf{link}$ to obtain a dendrogram $\mathsf{den}_i$.
  \If{$k_i$ is specified}
      \State Cut $\mathsf{den}_i$ at $k_i$ clusters to obtain $\mathcal{C}_i=\{C_{i1},\dots,C_{ik_i}\}$.
  \Else
      \For{$k=1$ to $|\mathcal{S}_i|$}
          \State Cut $\mathsf{den}_i$ at $k$ clusters to obtain $\mathcal{C}_i^{(k)}$.
          \State Temporarily assign stages according to $\mathcal{C}_i^{(k)}$ and \emph{refit} by pooling counts within each cluster and recomputing $\hat{\theta}$.
          \State Compute the global model score $S^{(k)}$ (e.g., BIC of the fitted staged tree).
      \EndFor
      \State Set $\mathcal{C}_i \gets \mathcal{C}_i^{(k_i^*)}$ where $k_i^*=\arg\max_k S^{(k)}$.
  \EndIf
  \State Update the function $\kappa$ (and thus $\eta$) in Definition~\ref{def:x-compatible-staged} by assigning the same stage label to all vertices in each cluster $C_{ij}\in\mathcal{C}_i$, and refit $\hat{\theta}$ for $X_i$ by pooling counts within clusters.
\EndFor
\State \textbf{Output:} an $X$-compatible staged event tree $(T,\kappa,\eta)$ and fitted probabilities $\hat{\theta}$.
\end{algorithmic}
\end{algorithm}

\subsubsection{The clustering algorithm}

We recover the stage structure by casting stage assignment as clustering of conditional probability vectors. For each context (internal vertex), the empirical conditional probabilities are computed and the resulting vector is treated as a point on the simplex. Hierarchical agglomerative clustering, equipped with a chosen distance or divergence (see Section~\ref{section:2}), groups contexts into stages, yielding parsimonious partitions while preserving the model’s probabilistic meaning.

Algorithm~\ref{alg:hc-stages}
is implemented in the \texttt{stagedtrees}  R package~\cite{JSSv102i06} %of the R statistical software~\cite{Rsoftware} 
and addresses both cases
with or without a predefined number of stages.
%\st{In this work, the method has been adapted to handle data lying on the simplex.
%When the number of stages is predetermined, the hierarchical clustering procedure proceeds by cutting the dendrogram at the specified number of clusters, thereby forming the exact number of stages $k_j$ for the variable $X_j$. 
%In contrast, if the number of stages is not known a priori, a fully data-driven model selection strategy is employed and multiple clustering configurations are systematically evaluated. For each candidate partition of the non-leaf vertices into stages, the corresponding staged event tree is refitted and its goodness-of-fit is assessed using a model selection criterion, typically the BIC. 
%The partition $\mathcal{C}_j^{(k_j^*)}$ that yields the highest  score is retained as the optimal stage configuration. }
The hierarchical clustering--based algorithm learns the stage structure of a staged event tree from an observed dataset 
$\mathcal{D}$ on a discrete random vector $X$. 
From $\mathcal{D}$, an initial saturated $X$-compatible event tree $T = (V, E)$ is constructed, where each internal vertex 
$v \in \mathring{V}$ (non-leaf node) %represents a distinct situation and 
is assigned a unique stage label $\kappa(v)$.

For each variable $X_i$ ($i = 2, \dots, p$), the algorithm considers the set of non-leaf nodes 
$\mathcal{S}_i$ at depth $i-1$ and  for each vertex $v \in \mathcal{S}_i$ estimates
the empirical conditional probability vector labelling its outgoind edges, 
$\hat{\theta}_v \in \Delta_{|\mathcal{X}_i|-1}$.
Pairwise similarities between these conditional distributions are quantified using a symmetric dissimilarity 
measure $d(\hat{\theta}_v, \hat{\theta}_w)$. In Section~\ref{sec:simulation} we use the Fisher, Jensen-Shannon, Kaniadakis, Total KL, Total Variation, Hellinger  measures. 
%, such as the Hellinger, total variation, Jensen--Shannon, or Kaniadakis divergence. 
The resulting dissimilarity matrix
\[
M_i = [\, d(\hat{\theta}_v, \hat{\theta}_w) \,]_{v,w \in \mathcal{S}_i}
\]
is used as input to a hierarchical clustering procedure with linkage $\mathsf{link}$. In Section~\ref{sec:simulation} we used the Average, Complete, McQuitty and Ward.D2 linkages.

If the number of stages $k_i$ is predefined, the dendrogram is cut at $k_i$ clusters, 
yielding $\mathcal{C}_i = \{C_{i1}, \dots, C_{ik_i}\}$. 
Otherwise, a fully data-driven model selection strategy is employed and multiple clustering configurations $\mathcal{C}_i^{(k)}$ are systematically evaluated, refitting the staged tree by pooling sufficient statistics. Its goodness-of-fit is assessed using a model selection criterion, typically the BIC
\[
k_i^* = \arg\max_k S(\mathcal{C}_i^{(k)}).
\]
The partition $\mathcal{C}_j^{(k_j^*)}$ that yields the highest  score is retained as the optimal stage configuration and the stage assignment $\kappa$ 
and edge labeling $\eta$ are updated. 
%Situations within the same cluster share a common parameter vector $\hat{\theta}$, ensuring statistical coherence.

Iterating this procedure for all $i = 2, \dots, p$ yields a fitted 
$X$-compatible staged event tree $(T, \kappa, \eta, \hat{\theta})$ that captures 
the essential context-specific conditional independence structured from the data.

\subsubsection{Clustering methods}

To ensure that the clustering procedure in Algorithm~\ref{alg:hc-stages} is robust, geometrically consistent, and computationally efficient, we employ four standard hierarchical linkage strategies: average, complete, McQuitty, and Ward.D2. These methods differ in how the distance between clusters is updated at each agglomeration step.
They are described in Table~\ref{tab:clustering_methods}, where each cluster \( \mathcal{C}_i \) contains stage-specific probability vectors representing conditional distributions for a variable \(X_j\). The dissimilarity \( d^{\mathrm{sym}} (p,q)\) between two probability density functions \(p\) and \(q\) is computed using one of the divergence or distance measures introduced in Section~\ref{section:2}.

\begin{table}
\small
\centering
\renewcommand{\arraystretch}{1.2}
\caption{Hierarchical clustering linkages used in Section~\ref{sec:simulation}. 
Here \(\mathcal{C}_1, \mathcal{C}_2, \mathcal{C}_3\) denote clusters of probability vectors \(p \in \Delta_{s-1}\);
\(d(p,q)\) is a dissimilarity between individual distributions; 
\(\bar p_i = |\mathcal{C}_i|^{-1}\sum_{p\in \mathcal{C}_i} p\) is the centroid of cluster \(\mathcal{C}_i\);
and \(\|\cdot\|\) is the Euclidean norm.}
\label{tab:clustering_methods}
\scalebox{0.9}{
\begin{tabular}{l l}
\toprule
Clustering method & Formula \\
\midrule
Average linkage &
$\displaystyle D_A(\mathcal{C}_1, \mathcal{C}_2)
= \frac{1}{|\mathcal{C}_1||\mathcal{C}_2|} 
\sum_{p \in \mathcal{C}_1}\sum_{q \in \mathcal{C}_2} d(p,q)$ \\

Complete linkage &
$\displaystyle D_C(\mathcal{C}_1, \mathcal{C}_2)
= \max_{p \in \mathcal{C}_1,\; q \in \mathcal{C}_2} d(p,q)$ \\

McQuitty (WPGMA) &
$\displaystyle D_M(\mathcal{C}_3, \mathcal{C}_1\!\cup\! \mathcal{C}_2)
= \tfrac{1}{2}\!\left[D(\mathcal{C}_3,\mathcal{C}_1)+D(\mathcal{C}_3,\mathcal{C}_2)\right]$ \\

Ward’s minimum variance (Ward.D2) &
$\displaystyle D_W(\mathcal{C}_1, \mathcal{C}_2)
= \frac{|\mathcal{C}_1||\mathcal{C}_2|}{|\mathcal{C}_1|+|\mathcal{C}_2|}
\,\|\bar{p}_1-\bar{p}_2\|^2$ \\
\bottomrule
\end{tabular}}
\end{table}

The \textit{average linkage} method defines the distance between two clusters as the mean of all pairwise dissimilarities, providing a balance between chaining effects and excessive compactness~\citep{rr1958statiscal}. 
The \textit{complete linkage} method instead considers the maximum pairwise dissimilarity, producing compact clusters when internal similarity is high. 
The \textit{McQuitty} (or WPGMA) method updates distances by assigning equal weight to the merged clusters, improving computational efficiency and avoiding repeated pairwise computations~\citep{mcquitty1966similarity}. 
Finally, \textit{Ward’s minimum variance} (Ward.D2) linkage minimizes the increase in within-cluster variance after each merge, leading to compact and well-separated clusters that often perform well in probability-based settings~\citep{ward1963hierarchical}. 
In practice, this method is widely used with various dissimilarity measures, as implemented for instance in the \texttt{hclust} function of R. 
Strictly speaking, however, Ward’s criterion is derived under squared Euclidean geometry, where the increase in within-cluster sum of squares can be expressed analytically~\citep{murtagh2014ward}. 
When applied to non-Euclidean dissimilarities, the procedure remains a valid hierarchical heuristic and often yields meaningful results, although the link with a formal variance-minimization objective no longer holds exactly.

%%%%%%%%%%%%%%%%%%%%%%%%%%%%
%%%%%%%%%%%%%%%%%%%%%%%%%%%%
%%%%%%%%%%%%%%%%%%%%%%%%%%%%

\section{Simulation Experiments}
\label{sec:simulation}

% We describe now the main simulation study to asses
% the performances of various hierarchical clustering techniques coupled with 
% different distances or divergences on the simplex. 

% We carry out an extensive simulation study to assess the efficacy of divergence-based hierarchical clustering techniques applied to synthetic datasets generated from random staged event tree models.
We carry out an extensive simulation study aimed at thoroughly assessing the effectiveness of divergence-based hierarchical clustering methods when utilized on simplex data generated from staged event tree models. In this investigation, we create synthetic datasets by simulating random staged event trees whose configurations and associated conditional probabilities represent points within the probability simplex. Each simulated tree results in a particular division of the simplex into regions that align with various stages, featuring context-specific conditional distributions that differ throughout the tree. By sampling data from these models, we gather  observations that exhibit realistic variability and context-dependent relationships. We then implement multiple hierarchical clustering techniques, combining them with various distance or divergence measures applicable to the simplex, to uncover the underlying stage groupings. The study systematically investigates the ability of these clustering approaches to identify the true staging arrangement under different conditions, including variations in sample size, tree complexity, and levels of noise.

%%%%%%%%%%%%%%%%%%%%%%%%%%%%
%%%%%%%%%%%%%%%%%%%%%%%%%%%%
%%%%%%%%%%%%%%%%%%%%%%%%%%%%
\subsection{Data Generation Process}\label{sec:syntheticmodelsdatasets}

We generate random staged event tree models  for a $p$-dimensional binary random vector with 
\( p \in \{5, 7, 9, 11\} \)
by defining random stages' structures in one 
of the three following ways: starting from a  saturated staged event tree model and

\begin{enumerate}
    \item randomly joining situations into stages at the same depth with a given probability $q \in \{0.5, 0.9\}$    
%    {\blue The parameter $q$  how we control the merging of situations (non-terminal nodes) in the same stage e.g each pair of situations $s_i,s_j$ we merge them to the same stage with probability $q$. for $q = 0 $ means we have fully saturated with no merging, $q = 0.5$ means moderate merging with medium number of stages, and lastly $q = 0.9$ means we have strong merging with small number of stages. It cannot be x-compatible because merging may happens across the contexts that correspond to different variable combinations. So, in this case $q$ controls random merging without enforcing conditional conditional structure.}
    \item or randomly splitting the situations at each depth into $k_0 = 2$ stages. 
%    {\blue The parameter $k$ controls the number of distinct stages at each level. Here we set $k_0 = 2 $ which means that all situations at each level are split into exactly two stages and shares a conditional probability distribution. Here $k_0$  defines exact number of stages that we want. This is exactly the x-comtible stage tree. In this case we can say the splitting is determinisitic while in first case we can say its probabilistic.}
\end{enumerate}

% This gives three possible methods to generate the stage structure.
The simulated staged event tree models are necessarily $X$-compatible, and those obtained with the second method have two stages at each depth with probability one. In the first method, the parameter $q$ controls the merging: $q = 0.5$ means moderate merging and $q = 0.9$ means strong merging. Once the  stages are generated,
conditional probabilities for each stage are randomly 
%\st{generated}
assigned through the \texttt{random\_sevt} function of the \texttt{stagedtrees} R-package with the default options~\cite{JSSv102i06}.

 Across all combinations of $p \in \{5, 7, 9, 11\} $, the three stage-generation methods, and  \( N \in \{2^7, 2^9, 2^{11}, 2^{13}\}\), a staged tree model is generated and a synthetic dataset from that staged tree model is simulated. One hundred independent replications were conducted for each combination, leading to a total of 4,800 synthetic datasets and  staged event tree models.

%one hundred independent replications were conducted {\blue generating a staged tree model and a dataset } \st{per setting}, leading to a total of 4,800 synthetic datasets and {\blue staged event tree models}. 
%For each {\blue $p$, for each of the three methods to generated the stages' structure, for each $N$}, we conduct $100$ independent replications to ensure a robust statistical assessment {\blue for a total of $4,800$ synthetic datasets}.  

% {\blue Once the stage assignment is defined i.e \(\kappa\), for each stage \( S  \in \mathcal{C}\) with outcome \( x_i \in \mathcal{X}_i\), the probability vector is drawn on the probability simplex. }\\

\subsection{Model Fitting}
\label{sec:model-fitting}
 
For each of the 4,800 synthetic datasets,
48 staged event tree models are fitted using  the \texttt{stage \_hclust} function 
with four different linkage methods (complete, average, Ward and McQuitty’s) and the six metrics in Tables~\ref{tab:distance_functions} and \ref{tab:divergence_functions}.
The parameter $k$  of the \texttt{stages\_hclust} function is either equal to two or NA.
This gives a total of 4,800$\times$48 staged tree models to be compared with reference baseline models constructed as follows.

%For each of the 4,800 synthetic datasets, a staged event tree model was fitted using hierarchical clustering of empirical probabilities as described in Section~\ref{sec:sevtHC}.
% Specifically, for every sample size $N$, we first fit a saturated staged event tree.  \st{Hierarchical clustering of the stages was then carried out }using the \texttt{stages\_hclust} function and employing different linkage methods (complete, average, Ward’s, and McQuitty’s) and the six metrics in Tables~\ref{tab:distance_functions} and \ref{tab:divergence_functions}. 
%This gives a total of 4,800$\times$24 staged tree models {\blue to be compared with reference baseline models constructed as follow.}

The first reference model is the data generating process by which the dataset is simulated according to Section~\ref{sec:syntheticmodelsdatasets}. 
The second reference model is the full saturated model, computed with the \texttt{full} function.
Finally, for datasets that have five or seven variables, a { reference} staged event tree model was fitted using an agglomerative hierarchical clustering algorithm to maximise the BIC score.  
This is a greedy search method, implemented in the \texttt{stages\_bhc} function, and is computationally unfeasible for  datasets with $p=9,11$.

% {\blue The staged event trees from which data are generated were considered as reference modes. }In addition, for datasets that have \st{at most 7} {\blue five or seven} variables, a { reference} staged event tree model was fitted using an agglomerative hierarchical clustering approach to maximise the BIC score. {\blue This is a greedy search method, implemented in the \texttt{stages\_bhc} function, and is computationally unfeasible for  datasets with $p=9,11$.}{\blue Furthermore, for all datasets {\red (or only for p=9,11?)} the full saturated model, computed with the \texttt{full} function, was considered as the { reference} staged event tree model.This gives two reference models for each datasets with $5,7$. {\red (why two reference models for $p=5,7$? How are they used?)}}

\subsection{Evaluation Measures}
For model comparisons we considered two different indices: the relative BIC and the relative Hamming distance. 

\subsubsection{Bayesian Information Criterion and Relative BIC}
The Bayesian Information Criterion (BIC) is a commonly employed method for model selection that balances good fit and model complexity. For a model $\mathcal M$ it is defined as 
\begin{equation*}  
\text{BIC}(\mathcal M) = -2 \log L(\hat{\theta}) + k \log N 
\end{equation*}
where  $L$ is the model likelihood function, \( \hat{\theta} \) the maximum likelihood estimates of the  model parameters, \( k \) the number of free parameters, and \( N \) for the sample size.
As for staged trees the model parameters are the stage conditional probabilities, an increase in the number of stages heightens the model’s complexity, and the BIC method penalizes this complexity unless it is offset by a considerable improvement in likelihood. 

BIC can be interpreted from both Bayesian and information-theoretic perspectives. From a Bayesian viewpoint, with a uniform prior across the model set \( \{ \mathcal{M}_1, \dots, \mathcal{M}_m \} \), the posterior probability of a model is roughly proportional to \( \exp(\text{BIC}(\mathcal{M}_i)) \). Hence, opting for the model with the highest BIC is similar to selecting the model with the greatest posterior probability. From an information-theoretic angle, BIC corresponds with the principle of Minimum Description Length, prioritizing models that offer the most succinct representation of the data \cite{schwarz1978estimating, wasserman2013}.

To evaluate the extent to which a model $\mathcal{M}$ diverges from a robust reference baseline $\mathcal{M}_{\text{B}}$, we used 
 the relative BIC difference defined as

%{\blue  It is useful to evaluate the extent to which a model $\mathcal{M}$ diverges from a robust reference baseline $\mathcal{M}_{\text{B}}$.} As model divergence we use the relative BIC difference defined as
%It is useful to evaluate the extent to which a model diverges from a robust reference baseline.
%As reference baseline we considered the BHC model {\blue questo e' il modello ottenuto con l'agglomerative hierarchical clustering altorithm} in Section~\ref{sec:model-fitting} for $p=7$ and a total of 1,200 synthetic datasets. 
%As model divergence we use the relative BIC difference between a specific model $\mathcal{M}$  and the reference model $\mathcal{M}_{\text{BHC}}$:}
%\[ 
%\Delta_{BIC}(\mathcal{M}_i , \mathcal{M}_{\text{BHC}}) = \frac{1}{100} \sum_{j=1}^{100} \frac{\text{BIC}(\mathcal{M}_i^{(j)}) - \text{BIC}(\mathcal{M}_{\text{BHC}}^{(j)})}{|\text{BIC}(\mathcal{M}_{\text{BHC}}^{(j)})|},  
%\] 
\[
\Delta_{BIC}(\mathcal{M} , \mathcal{M}_{\text{B}}) = \frac{\text{BIC}(\mathcal{M}) - \text{BIC}(\mathcal{M}_{\text{B}})}{|\text{BIC}(\mathcal{M}_{\text{B}})|} . 
\] 
% where the average is computed over the hundred replications.
%for each $N$ \st{denoted by} \( j \) , and is average of relative differences. where the summation is taken over multiple replications \( j = 1, \dots, n \).  
A value of \( \Delta_{\mathrm{BIC}} \approx 0 \) indicates that  $\mathcal M$ achieves a fit comparable to the benchmark model, while larger positive values reflect a poorer fit. 
% {\blue While the BIC takes values from plus to minus infinity, the relative BIC where $\mathcal B$ is better fitting than $\mathcal B$ is negative.}
Using a relative rather than an absolute difference normalizes the measure and enables 
comparability across experiments with different sample sizes or model dimensions, a property that BIC lacks as it increases with the sample size.
 
% A \( \Delta_{BIC} \) value that is near zero suggests that the model performs similarly to the BHC model \st{with respect to BIC}, whereas larger positive \( \Delta_{BIC} \) \st{values} indicates a less satisfactory fit in comparison to the BHC benchmark. Notably, opting for a relative difference instead of an absolute one standardizes the comparison and facilitates more equitable interpretation across datasets with differing scales or complexities.  

\subsubsection{ Hamming Distance and Relative Hamming Distance}

%\st{This} {\blue The \( \Delta_{BIC} \) } diagnostic enhances other model evaluation metrics by basing assessments on a reference model recognized for its equilibrium of complexity and fit. \st{In the framework of } 
In a staged event tree, every node represents a variable context and is assigned a stage that reflects its conditional probability distribution. The Hamming distance between two staged tree models assesses the quantity of node-level stage assignments that vary between the reference model and the estimated model. This metric serves as a clear and meaningful way to measure structural accuracy, particularly in scenarios related to stage clustering or model simplification. 
The Hamming distance between two models is calculated with the 
{\texttt{compare\_stages}} function of the {\texttt{stagedtrees}} R-package, which relies on the {\texttt{clue}} package.
A smaller Hamming distance suggests a closer alignment in structural configuration between the two models, while a larger value signifies a considerable divergence in the probability structure of the two models.
Let \( \kappa_1 \) and \( \kappa_2 \) denote the stage function in Definition~\ref{def:x-compatible-staged} for the two models 
\(\mathcal{M}_1\) and \(\mathcal{M}_2\), respectively. 
Their Hamming distance is defined as
\[
\text{HD}(\mathcal{M}_1, \mathcal{M}_2)
= \sum_{v \in \mathring{V}} \mathbf{1}\big[\kappa_1(v) \neq \kappa_2(v)\big] 
\]
where $\mathbf{1}$ stands for indicator function.

The relative Hamming distance for a model $\mathcal M$ with respect to two reference models $\mathcal M_O$  and $\mathcal M_B$ is defined as
\[
\Delta_{\text{HD}} (\mathcal{M} ,  \mathcal{M}_B ) = 
\frac{ \text{HD}(\mathcal{M}, \mathcal{M}_O) - \text{HD}( \mathcal{M}_B, \mathcal{M}_O )
}{\text{HD}( \mathcal{M}_B, \mathcal{M}_O )}
\]
where $\mathcal M_O$ is the generating data process, that is the original staged tree model from which the synthetic dataset, on which $\mathcal{M}$ was fitted, was drawn. The relative Hamming distance quantifies how close the learned model $\mathcal{M}$ is to the true data-generating process $\mathcal{M}_O$, expressed relative to a baseline model $\mathcal{M}_B$. The numerator measures the improvement (or deterioration) in structural accuracy of $\mathcal{M}$ compared to the baseline, while the denominator normalizes this difference so that values are interpretable across settings.

\subsection{Median Classification Accuracy}
In summary, in Subsection~\ref{sec:syntheticmodelsdatasets} 4,800 synthetic datasets were generated by varying the number of variables $p\in \{5,7,9,11\}$, the generating methods $m$ (given by either $q\in \{0.5,0.9\}$ or $k_0=2$), the sample size $N\in \{ 2^7,2^9,2^{11},2^{13} \}$ and hundred independent replicates $r$ for each combination of $(p, m, N)$. For each dataset, in Subsection~\ref{sec:model-fitting} we fitted 48 clustering based staged tree models, one for each combination of four linkage functions (indicated with $l$), six metrics (indicated with $d$) and two possible number of stages for each variables (indicated with $k$). Furthermore, we considered two reference models: the original model $\mathcal M_O$ and the full model $\mathcal M_F$, and third reference model $\mathcal M_{BHC}$ is estimated with the backward hill climbing algorithm.

For each of the 230,400 fitted models, we computed six indices of goodness-of-fit: the relative BIC deviation and the relative Hamming distance with respect to the two possible reference models, namely $\mathcal M_F$ and $\mathcal M_{BHC} $. 
This gives four values for each combination of $(p, m, N, r; l, d, k)$.
Finally, across the hundred replicates $r$ we computed the median values of each of the four goodness-of-fit indices. The median, rather than the mean, was selected to provide a more robust measure of central tendency that is less influenced by extreme values or outlier runs, which can occasionally occur in stochastic simulations. Using the median ensures that the reported goodness-of-fit indices better reflect the typical performance across replications rather than being skewed by a few atypical cases.

These $4 \times 2,304$ median values are used in Section~\ref{ref:resultsSimulation} to assess the effectiveness of divergence-based hierarchical clustering methods by determining how closely each fitted model aligns with the original stage structure from which data were simulated in comparison to reference baseline robust models.

%{\red  \label{Goodness} This section focuses on assessing the goodness of fit, highlighting the differences among clustering methods, and identifying the distances that perform most effectively. 
%For each of the 4,800$\times$24 staged tree models obtained in Section~\ref{sec:model-fitting}, we determine how closely it aligns with the original stage structure from which data were simulated. We employ three different methods to evaluate the discrepancy between the true model and the estimated model: the BIC, the relative BIC and the Hamming distance.}

\subsection{Results of the Simulation} \label{ref:resultsSimulation}

The results of simulations can be summarised in twenty four panels. Each panel corresponds to a fixed value of 
$(p,m,k)$.  
In each panel the median relative BIC and HD with respect to the full reference model $\mathcal M_{F}$ are plotted for each combination of the linkage methods ($l$, in columns) and of the distances ($d$, given by the different colors). Furthermore, for $p=5$ and $7$ black dots give the 
median relative BIC and HD with respect to the BHC reference model $\mathcal M_{BHC}$. 
% See Figures~\ref{} ..... 

% Another way to present the simulation results is as in Figures~\ref{} ... where
% in each panel $(N,m, k)$ are fixed.

% NEXT COMES THE COMMENTS OF THE FIGURES. 

% - obviously $k=NA$ is better than $k=2$

% - $BIC_{Full}$ is always positive from zero to 121,608

% - $BIC_{BHC}$ ranges from zero to 74,010

% - $BIC_{fitted}$ ranges for different  

\begin{figure}[H]
    \centering
    \includegraphics[width=\textwidth]{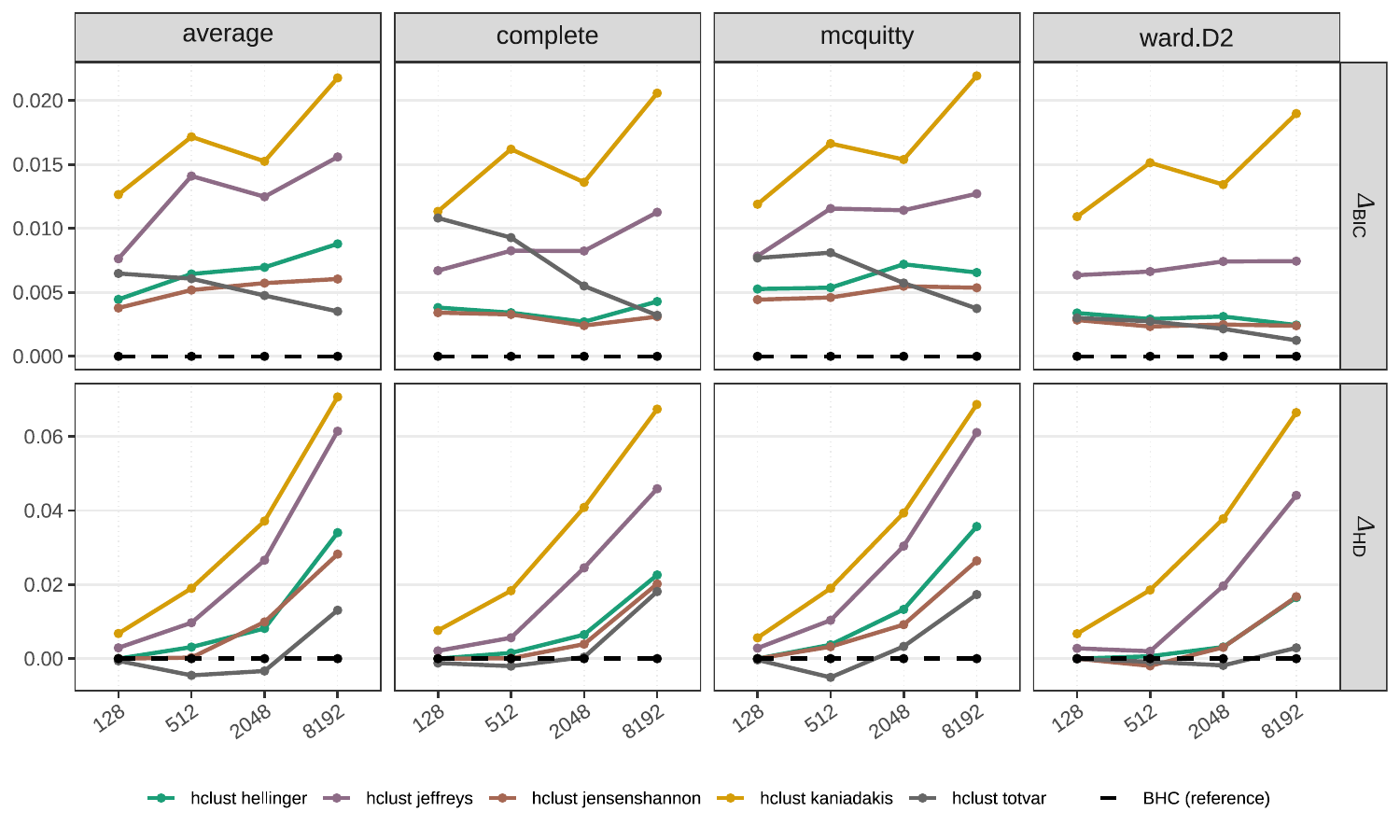}
    \caption{Comparison of  \(\Delta_{BIC}\) and \(\Delta_{HD}\) across divergence-based staged event tree classifiers for $p = 11$, under the following configurations: 
    \( k_{0} ,  k = NA, q = 0.9.\)
    }
    \label{fig:divergence_methods}
\end{figure}

 Figure~\ref{fig:divergence_methods}evaluates the performance of five divergence-based hierarchical clustering classifiers for staged event trees i.e Hellinger, Jeffreys, Jensen-Shannon, Kaniadakis, and total variation against the BHC reference (marked by the black dashed line at $0$). It provides information on the relative BIC deviation (top row, $\Delta_{\mathrm{BIC}}$) and relative Hamming distance (bottom row, $\Delta_{\mathrm{HD}}$) as sample sizes increase, specifically for $n\in\{128,512,2048,8192\}$ and four different linkage methods (average, complete, McQuitty, Ward.D2) while maintaining $p=11$, $k_0$ fixed, $k=\mathrm{NA}$, and $q=0.9$. For all linkage methods examined, \texttt{Ward.D2} consistently produces the lowest deviations, with {total variation} (in gray) and Jensen-Shannon (in brown) often being the closest to the BHC benchmark for both model fit and structural recovery, which suggests they maintain stable performance as $n$ increases. In contrast, Kaniadakis (in gold) displays the most significant deviation from the reference across all panels its $\Delta_{\mathrm{BIC}}$ remains the highest, and its $\Delta_{\mathrm{HD}}$ sharply escalates with larger sample sizes while Jeffreys (depicted in purple) typically shows the second-highest level of divergence, particularly in cases of structural disagreement. In summary, the figure demonstrates that utilizing a strong divergence measure most notably total variation in conjunction with an "optimizing" linkage such as \texttt{Ward.D2} offers the most dependable balance between model accuracy (indicated by a small $\Delta_{\mathrm{BIC}}$) and structural integrity (reflected by a small $\Delta_{\mathrm{HD}}$) within this $p=11$ framework.

\begin{figure}[H]
    \centering
    \includegraphics[width=\textwidth]{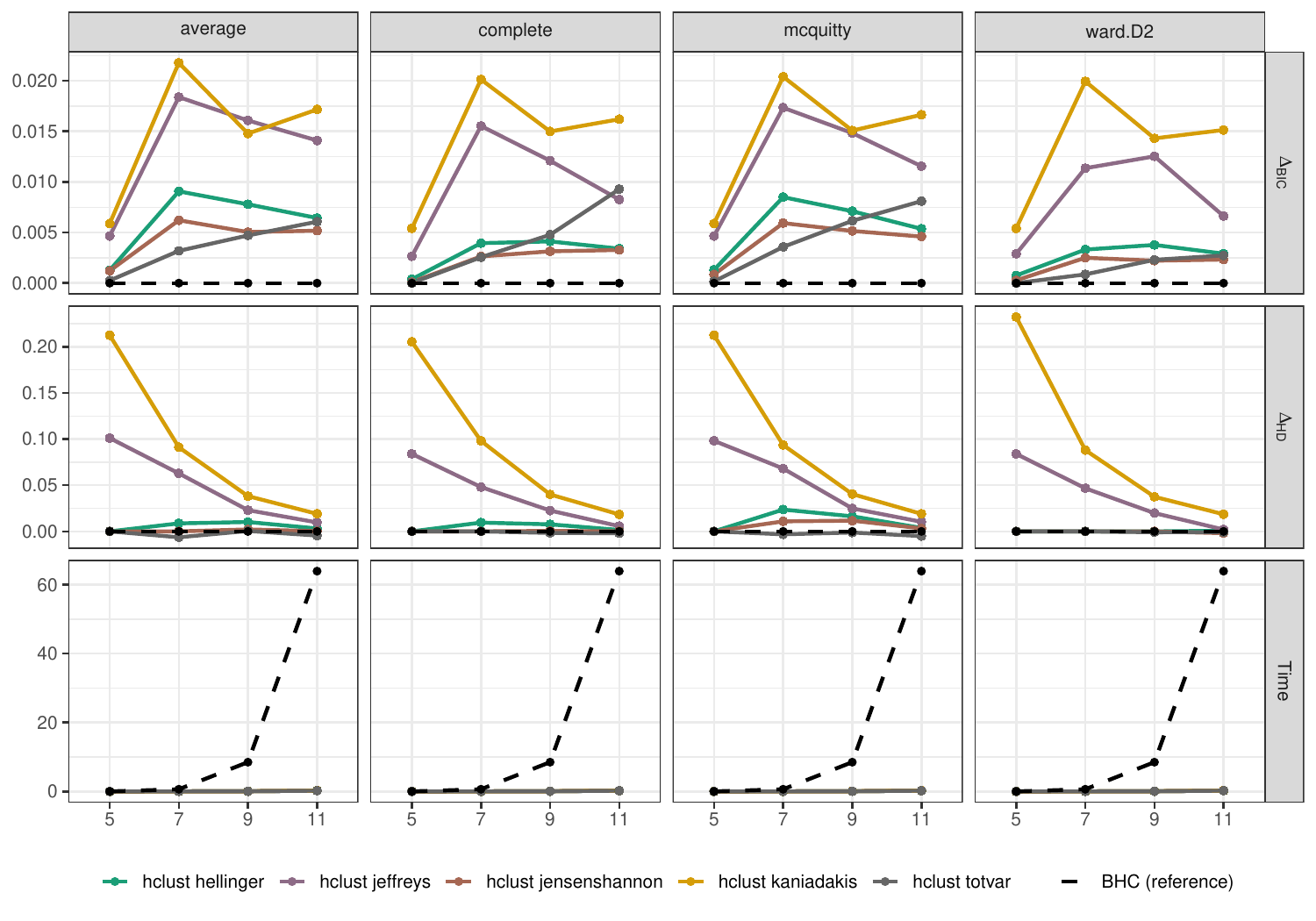}
    \caption{Comparison of $\Delta_{\mathrm{BIC}}$ and $\Delta_{\mathrm{HD}}$ and time  across divergence-based staged event tree classifiers for sample sizes $512$ with \(k_{0}, k  = NA, q = 0.9\)}
    \label{fig:dive_meth_time}
\end{figure}

  Figure~\ref{fig:dive_meth_time} illustrates the effects of increasing the number of variables used to construct the model (the values on the $x$-axis: $5,7,9,11$). You can view moving from left to right as enlarging the problem: the model needs to take into account more information simultaneously. The four columns represent different methods of forming clusters (average, complete, McQuitty, and Ward.D2), while the three rows reflect (i) the performance of the fitted model relative to a reference method (top row: $\Delta_{\mathrm{BIC}}$), (ii) how distinct the recovered tree structure is from the reference (middle row: $\Delta_{\mathrm{HD}}$), and (iii) the computing time necessary to execute the method (bottom row: Time in seconds). The main practical takeaway lies  in the bottom row: as the number of variables increases, the dashed black reference method (BHC) experiences a significant slowdown, sharply rising to about tens of seconds at $11$ variables, while the colored divergence-based hierarchical clustering methods maintain a flat runtime close to zero seconds across all scenarios. This indicates that while increasing the number of variables complicates the task, our divergence-based methods exhibit a much more gradual increase in runtime compared to the reference method, all while achieving comparable accuracy and structure recovery (as shown by small $\Delta_{\mathrm{BIC}}$ and $\Delta_{\mathrm{HD}}$).

In our simulation, we create staged event trees through two random processes: (i) random stage formation, in which circumstances at the same depth are randomly grouped into common stages with a specified probability $q \in \{0.5, 0.9\}$; and (ii) random stage splitting, where each depth sees situations randomly divided into a fixed number of stages, which in this case is $k_0 = 2$. The primary results presented in the main text, shown in Figures~\ref{fig:divergence_methods} and \ref{fig:dive_meth_time}, pertain to the high-staging scenario with $q=0.9$, while the corresponding plots (along with any other relevant configurations) can be found in the Supplementary Material for thoroughness.

\section{Application to classification tasks}

% Clustering methods to estimate the stages' structure have been employed to fit so-called naive staged event tree classifiers~\cite{carli2023new}.  

% We thus explore here how different distances affect the performances of naive staged event tree classifiers obtained employing a hierarchical clustering approach. 

%We assess the effectiveness of the classifiers under consideration on a set of benchmark datasets, employing a methodology akin to that used by \cite{carli2023new}. 

%% Table~\ref{tab:dataset_summary} presents a summary of the datasets, highlighting the normalized entropy of the class variable as a metric for class imbalance.

\begin{figure}[H]
    \centering
    \includegraphics[width=1\textwidth]{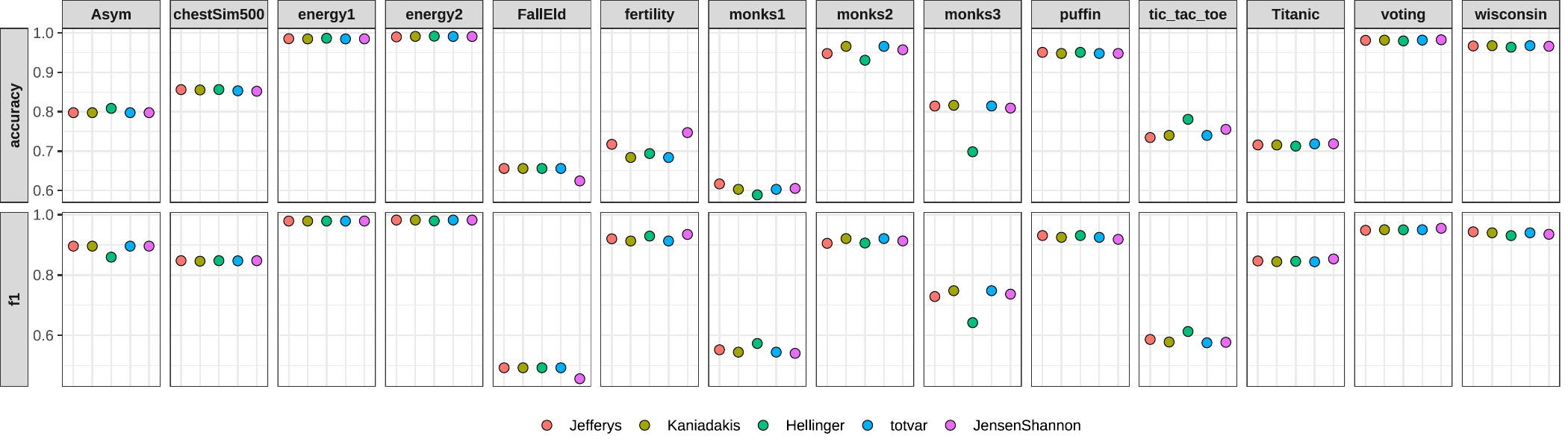}
    \caption{Accuracy and f1  for all the staged event tree classifiers utilizing various divergence measures $k = 2$}
    \label{fig:accuracy}
\end{figure}

Clustering methods to estimate the stage structure have been used to fit so-called naive staged event tree classifiers~\cite{carli2023new}. Here, we explore how different distance measures affect the performance of naive staged event tree classifiers obtained using a hierarchical clustering approach. We assess the effectiveness of the classifiers on a set of fourteen benchmark datasets, employing a methodology similar to that used by~\cite{carli2023new}. 
For each dataset, we conduct 10 iterations of an 80\%--20\% train-test split and present the median classification accuracies and median F1 score in Figures~\ref{fig:accuracy}). 
% Figures~\ref{fig:accuracy} compare the performance of staged event tree classifiers constructed using different divergence measures in combination with Ward.D2 hierarchical clustering with $k = 2 $. Divergene measures like Total Variation, Jefferys, Jensen-Shannon, and Hellinger exhibit more consistent and dependable performance, producing similar accuracy across a diverse array of datasets. This consistency underscores their appropriateness for general-purpose staged event tree construction when integrated with hierarchical clustering.

% Figure~\ref{fig:accuracy} displays the predictive performance of the divergence-based staged event tree classifiers across various benchmark datasets, with each small panel representing one specific dataset. 
Each pair of vertical panels in Figure~\ref{fig:accuracy} displays the predictive performance of five divergence-based staged event tree classifiers for a benchmark dataset. 
The upper section shows accuracy, defined as the ratio of correctly classified observations, while the lower section presents the $F_{1}$-score, which balances precision and recall, making it especially useful when class sizes are imbalanced. 
The colored points represent the five divergences (Jeffreys, Kaniadakis, Hellinger, total variation, and Jensen--Shannon);
points at the same height suggest that the choice of divergence has a minimal impact on predictive performance for the particular dataset. 
%The figure is arranged into two sections: the upper section shows accuracy, defined as the ratio of correctly classified observations, while the lower section presents the $F_{1}$-score, which balances precision and recall, making it especially useful when class sizes are imbalanced. Within each dataset panel, the colored points represent the five divergences (Jeffreys, Kaniadakis, Hellinger, total variation, and Jensen--Shannon); points that are grouped closely together suggest that the choice of divergence has a minimal impact on predictive performance for that particular dataset. 

%A notable trend is that numerous 
Five datasets are essentially ``solved" with all divergences, with both accuracy and $F_{1}$ scores approaching $1$ (for instance, energy1, energy2, \texttt{voting}, and wisconsin), indicating that the classifiers are very dependable when the signal is strong. For moderately challenging tasks (like Asym and chestSim500), the points again are at similar height, reflecting resilience to the choice of divergence and only slight variations in performance. The most difficult datasets (specifically FalEld, fertility, and monks1) show overall lower scores and a somewhat greater difference between divergences, with the $F_{1}$-row often highlighting more distinct variations than accuracy; this suggests that performance sensitive to class can decline also when overall accuracy stays moderate. In more challenging situations, total variation and Jensen-Shannon are often among the top-performing options, while Jeffreys and Kaniadakis usually follow closely behind, and Hellinger occasionally ranks a bit lower. To summarize, Figure~\ref{fig:accuracy} suggests that the divergence-based classifiers are largely stable across different divergences, but the choice of divergence becomes more critical in difficult datasets, where total variation and Jensen--Shannon yield strong predictive performance in both accuracy and $F_{1}$.

\section{Conclusion}
\label{con}

This paper introduced a framework that is both geometrically aware and computationally efficient for learning $X$-compatible staged event tree models, treating stage recovery as a hierarchical clustering challenge on the probability simplex. The main concept is to depict each situation (at a specific depth) by its empirical conditional probability vector and to measure the similarity between situations using well-defined simplex dissimilarities (distances/divergences). By creating level-wise dissimilarity matrices and integrating them with traditional linkage methods, we establish a clear family of potential stage partitions that can be chosen based on a well-known model-selection criterion (in this case, BIC). By doing this, we substitute purely score-driven greedy search methods with a modular pipeline that allows inspection, comparison, and tuning of its components (divergence, linkage, and cut selection) to align with the data-generating process.

Our simulation investigation revealed two interrelated insights. Firstly, from a statistical standpoint, the choice of divergence and linkage significantly affects both fit and structure recovery: across different sample sizes and configurations, \texttt{Ward.D2} consistently delivered the most reliable partitions, while Total Variation divergence (often closely trailed by Jensen-Shannon) frequently emerged as one of the top performers in terms of relative BIC deviation and relative Hamming distance to the BHC benchmark. Secondly, from a computational viewpoint, the divergence-based clustering methods scale considerably better than BHC as the number of variables escalates: although the runtime for BHC increases sharply (reaching several seconds in our higher-dimensional scenarios), the divergence-based methods maintain an almost constant runtime close to zero seconds, making them preferable for larger models, repeated fitting (such as cross-validation), and exploratory analysis. We enhanced the simulation findings by conducting a classification analysis on benchmark datasets, where the staged-tree classifiers consistently demonstrated high accuracy and $F_1$ scores; notably, the performance remained generally stable regardless of the divergence selection, although challenging datasets highlighted minor yet significant advantages for Total Variation and Jensen--Shannon, particularly when evaluated using the class-sensitive $F_1$ metric.

Collectively, these findings reinforce a key methodological insight: employing staged tree estimation benefits from honoring the inherent geometry of probability vectors, while a divergence-based hierarchical clustering approach offers a viable and interpretable pathway to stage learning that maintains robust statistical performance and significantly lowers computational demands. 
% \begin{figure}[H]
% \centering
% \subfloat[ Q=0.5, and \(\text{K}_{\text{true}}\)=NA]{\label{i}\includegraphics[width=.48\linewidth]{images/_Q0.5 Ktrue =NA.pdf}}\hfill
% \subfloat[ Q=NA, and \(\text{K}_{\text{true}}\)=2]{\label{j}\includegraphics[width=.48\linewidth]{images/_QNA Ktrue =2.pdf}}\par 

% \caption{BIC Score, Hamming Distance with different variations of Q, and \(\text{K}_{\text{true}}\) }
% \label{fig3}
% \end{figure}

% \begin{figure}[H]
% \centering
% \subfloat[ Q=0.5, and \(\text{K}_{\text{true}}\)=NA]{\label{k}\includegraphics[width=.48\linewidth]{images/Q0.5 Ktrue =NA.pdf}}\hfill
% \subfloat[ Q=NA, and \(\text{K}_{\text{true}}\)=2]{\label{l}\includegraphics[width=.48\linewidth]{images/QNA Ktrue =2.pdf}}\par 

% \caption{BIC Score, Hamming Distance with different variations of Q, and \(\text{K}_{\text{true}}\) }
% \label{fig4}
% \end{figure}

% \pagebreak

\section*{Declaration of competing interest}
The authors declare that they have no known competing financial interests or personal relationships that could have appeared to influence the work reported in this paper.

\section*{Acknowledgments}
This project has received funding from the European Union’s Horizon 2020 research and
innovation programme under Marie Skłodowska-Curie GA No. 101034449.

\section*{Data availability}
The data that support the findings of this study are openly available in Github at \url{https://github.com/stagedtrees/stagedtrees_clustering_new}.

\bibliographystyle{elsarticle-num} 
\bibliography{biblio}

%% Use figure environment to create figures
%% Refer following link for more details.
%% https://en.wikibooks.org/wiki/LaTeX/Floats,_Figures_and_Captions

\appendix
\section*{Appendix A Mathematical Notation and Definitions}
\begin{table}[H]
\centering
\caption{Summary of key mathematical notation used in the learning algorithm~\ref{alg:hc-stages}}
\begin{tabular}{ll}
\toprule
\textbf{Symbol} & \textbf{Meaning} \\
\midrule
$T = (V,E)$ & Event tree with vertices $V$ and edges $E$ \\
$\mathring{V}$ & Set of non-leaf (internal) vertices \\
$\kappa : \mathring{V} \to \mathcal{C}$ & Stage assignment function \\
$\eta : E \to \mathcal{C} \times \bigcup_i \mathcal{X}_i$ & Edge-labeling map \\
$\mathcal{S}_i$ & Set of situations at level $i-1$ \\
$\hat{\theta}_v$ & Empirical conditional probability vector at vertex $v$ \\
$d(\cdot, \cdot)$ & Dissimilarity or divergence on the simplex \\
$M_i$ & Dissimilarity matrix for situations at level $i$ \\
$k_i$ & Number of stages for variable $X_i$ \\
$S^{(k)}$ & Model selection score for $k$-cluster configuration \\
$k_i^*$ & Optimal number of stages maximizing $S^{(k)}$ \\
\bottomrule
\end{tabular}
\end{table}

\section*{Appendix B Benchmark Datasets Details}
\begin{table}[H]
\footnotesize
\centering
\caption{Summary of the 14 datasets used in the experimental study, including number of observations, number of variables, number of atomic events, and imbalance measure.}
\label{tab:dataset_summary}
\vspace{0.3em}

\begin{tabular}{lrrrr}
\toprule
Dataset        & \# Observations & \# Variables & \# Atomic Events & Imbalance \\
\midrule
Asym           & 1{,}000& 4  & 13    & 0.7 \\
chestSim500    & 500    & 8  & 27    & 0.546 \\
energy1        & 768    & 9  & 1{,}728    & 1.000 \\
energy2        & 768    & 9  & 1{,}728    & 1.000 \\
fallEld        & 5{,}000  & 4  & 64        & 0.888 \\
fertility      & 100    & 10 & 15{,}552   & 0.529 \\
monks1         & 432    & 7  & 864       & 1.000 \\
monks2         & 432    & 7  & 864       & 0.914 \\
monks3         & 432    & 7  & 864       & 0.998 \\
puffin         & 69     & 6  & 768       & 0.999 \\
ticTacToe      & 958    & 10 & 39{,}366   & 0.931 \\
titanic        & 2{,}201  & 4  & 32        & 0.908 \\
voting         & 435    & 17 & 131{,}072  & 0.997 \\
wisconsin      & 683    & 10 & 1{,}024    & 0.934 \\
\bottomrule
\end{tabular}
\end{table}

\section*{Appendix C: Supplementary Material and Code Availability}
Additional figures and supplementary results, along with the code used for simulations, model fitting, and plotting, are available in the project repository. This material provides further experimental settings and supporting analyses that complement the results presented in the main manuscript, and it enables full reproducibility of the computational workflow:
\url{https://github.com/stagedtrees/stagedtrees_clustering_new}.
\begin{figure}[H]
    \centering
    \includegraphics[width=\textwidth]{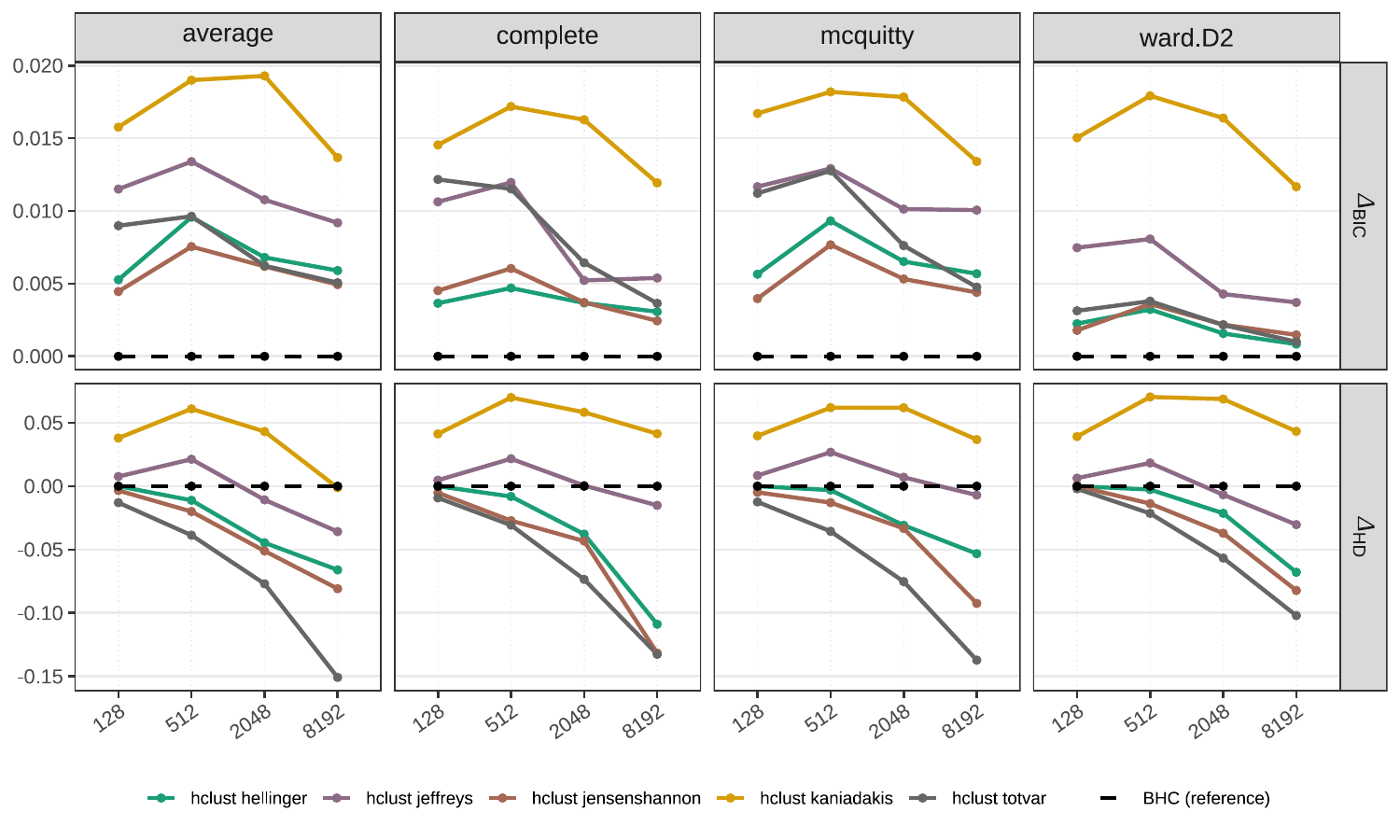}
    \caption{Comparison of  \(\Delta_{BIC}\) and \(\Delta_{HD}\) across divergence-based staged event tree classifiers for $p = 11$, under the following configurations: 
    \( k_{0} = 2,  k = NA, q = NA\)
    }
    \label{fig:divergence_methods_1}
\end{figure}

\begin{figure}[H]
    \centering
    \includegraphics[width=\textwidth]{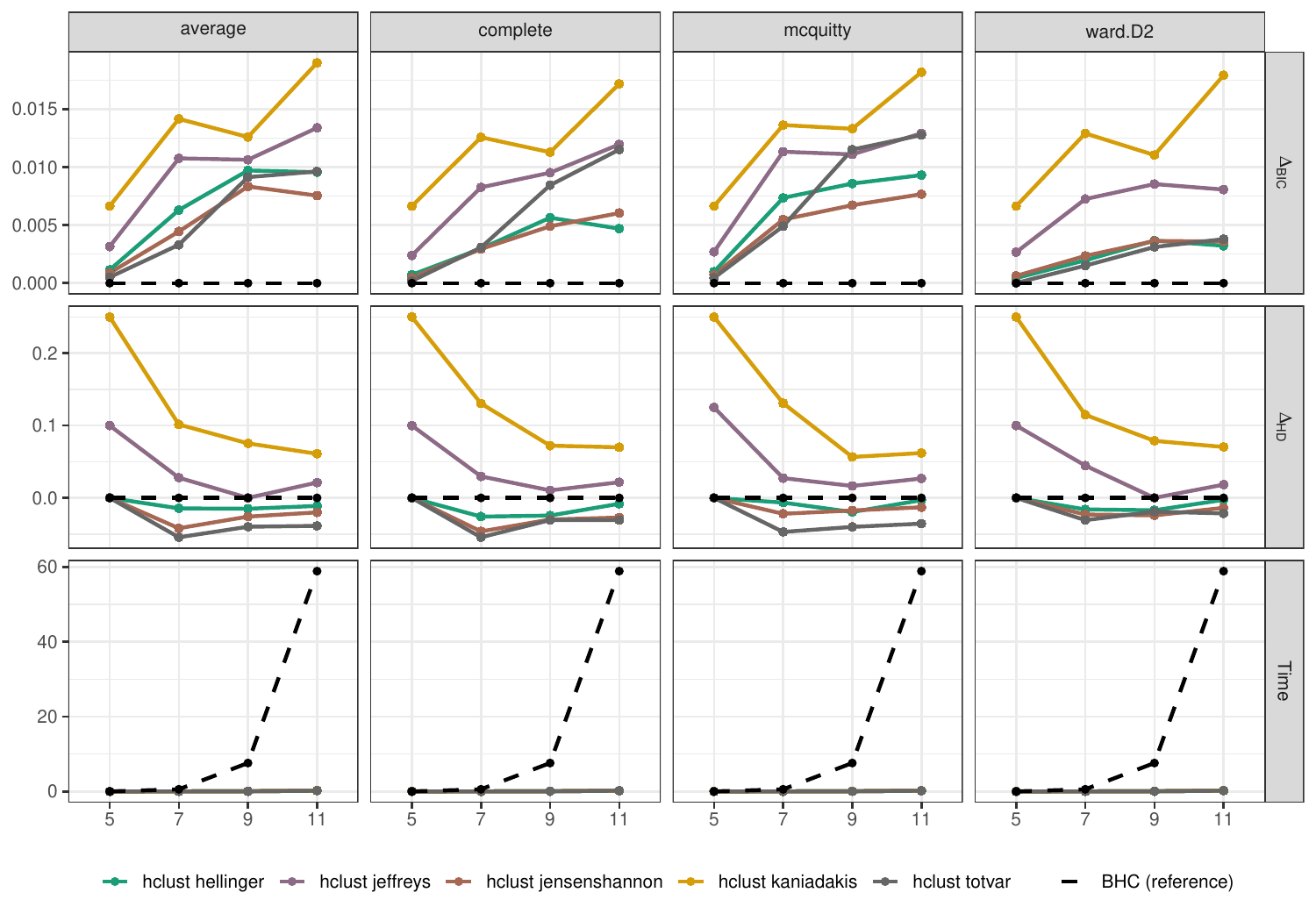}
    \caption{Comparison of $\Delta_{\mathrm{BIC}}$ and $\Delta_{\mathrm{HD}}$ and time  across divergence-based staged event tree classifiers for sample sizes $512$ with \(k_{0} = 2, k  = NA, q = NA\)}
    \label{fig:divergence_methods_2}
\end{figure}

%%       \cite{<label>} ==> [1]


\end{document}